%% file: main.tex
\documentclass{article}

\usepackage[nonatbib,final]{neurips_data_2023}

\usepackage[utf8]{inputenc} % allow utf-8 input
\usepackage[T1]{fontenc}    % use 8-bit T1 fonts
\usepackage{hyperref}       % hyperlinks
\usepackage{url}            % simple URL typesetting
\usepackage{booktabs}       % professional-quality tables
\usepackage{amsfonts}       % blackboard math symbols
\usepackage{nicefrac}       % compact symbols for 1/2, etc.
\usepackage{microtype}      % microtypography

\usepackage{dsfont}
\usepackage{pifont}
\usepackage{xspace}
\usepackage{caption}
\usepackage{colortbl}
\usepackage{multirow}
\usepackage{graphicx}
\usepackage{mdframed}
\usepackage[usenames,dvipsnames]{xcolor}

\newcommand{\bench}{\textsc{SpreadsheetBench}\xspace}

\title{\bench: Towards Challenging \\ Real World Spreadsheet Manipulation}

\author{%
  Zeyao Ma\textsuperscript{\normalfont{1,$\ast$}},
  Bohan Zhang\textsuperscript{\normalfont{1,$\ast$}},
  Jing Zhang\textsuperscript{\normalfont{1,$\dagger$}},
  Jifan Yu\textsuperscript{\normalfont{2}},
  Xiaokang Zhang\textsuperscript{\normalfont{1}},
  \\
  \textbf{
  Xiaohan Zhang\textsuperscript{\normalfont{3}},
  Sijia Luo\textsuperscript{\normalfont{1}},
  Xi Wang\textsuperscript{\normalfont{1}},
  Jie Tang\textsuperscript{\normalfont{2}}}
  \\
  \textsuperscript{1} Renmin University of China
  \textsuperscript{2} Tsinghua University
  \textsuperscript{3} Zhipu.AI \\
  \texttt{https://spreadsheetbench.github.io} \\
}

\begin{document}

\maketitle

\begingroup\def\thefootnote{$\ast$}\footnotetext{Equal Contribution.}\endgroup
\begingroup\def\thefootnote{$\dagger$}\footnotetext{Corresponding author.}\endgroup

\input{latex/0.abstract}

\section{Introduction}
\label{sec:introduction}
\input{latex/1.introduction}

\section{\bench}
\label{sec:benchmark}
\input{latex/2.benchmark}

\section{Experiments}
\label{sec:experiments}
\input{latex/3.experiments}

\section{Conclusion}
\label{sec:conclusion}
\input{latex/5.conclusion}

\begin{ack}
% Use unnumbered first-level headings for the acknowledgments. All acknowledgments
% Go at the end of the paper before the list of references. Moreover, you are required to declare
% funding (financial activities supporting the submitted work) and competing interests (related financial activities outside the submitted work).
% More information about this disclosure can be found at: \url{https://neurips.cc/Conferences/2023/PaperInformation/FundingDisclosure}.
% You can use the \texttt{ack} environment provided in the style file. As opposed to the main NeurIPS track, acknowledgements do not need to be hidden.
The authors would like to thank the annotators from Zhipu.AI who participated in the research and data collection process.
This work is supported by the National Key Research \& Develop Plan (2023YFF0725100) and National Natural Science Foundation of China (62322214, 62076245, U23A20299, 62072460, 62172424, 62276270).

\end{ack}

\bibliographystyle{ieeetr}
\bibliography{neurips_data_2024}

%%%%%%%%%%%%%%%%%%%%%%%%%%%%%%%%%%%%%%%%%%%%%%%%%%%%%%%%%%%%
\section*{Checklist}
\label{sec:checklist}
\input{latex/6.checklist}

%%%%%%%%%%%%%%%%%%%%%%%%%%%%%%%%%%%%%%%%%%%%%%%%%%%%%%%%%%%%

\clearpage

\appendix
\label{sec:appendix}
\input{latex/7.appendix}

\end{document}

%% file: latex/0.abstract.tex
\begin{abstract}

We introduce \bench, a challenging spreadsheet manipulation benchmark exclusively derived from real-world scenarios, designed to immerse current large language models (LLMs) in the actual workflow of spreadsheet users. 
Unlike existing benchmarks that rely on synthesized queries and simplified spreadsheet files, \bench is built from 912 real questions gathered from online Excel forums, which reflect the intricate needs of users. The associated spreadsheets from the forums contain a variety of tabular data such as multiple tables, non-standard relational tables, and abundant non-textual elements. Furthermore, we propose a more reliable evaluation metric akin to online judge platforms, where multiple spreadsheet files are created as test cases for each instruction, ensuring the evaluation of robust solutions capable of handling spreadsheets with varying values.
Our comprehensive evaluation of various LLMs under both single-round and multi-round inference settings reveals a substantial gap between the state-of-the-art (SOTA) models and human performance, highlighting the benchmark's difficulty.

\end{abstract}

%% file: latex/1.introduction.tex
% Background
Today, millions of users engage with spreadsheets across various fields, including management, marketing, and finance~\cite{rahman2020benchmarking,jackson2006excel_finance,winston2014excel_marketing}. Assisting users with spreadsheet manipulation can significantly reduce their workload, thereby enhancing productivity and efficiency~\cite{gulwani2011automating, cheung2016custodes}. 
Initially, program synthesis aided spreadsheet manipulation~\cite{gulwani2011automating-string, gulwani2012spreadsheet-man}, followed by deep learning methods for automated manipulation~\cite{chen2021spreadsheetcoder, chen2024auto-formula}.
More recently, leveraging the powerful capabilities of LLMs~\cite{achiam2023gpt-4}, spreadsheet agents such as SheetCopilot\cite{li2024sheetcopilot} and SheetAgent~\cite{chen2024sheetagent} have been developed to assist people in manipulating spreadsheets via natural language instructions. These agents employ LLMs to transform instructions into solutions, typically presented as executable code for spreadsheet operations.
% We anticipate that these advancements will substantially simplify the process of spreadsheet manipulation.

% Target
However, current spreadsheet-related agents~\cite{li2024sheetcopilot,chen2024sheetagent} are still far from being truly helpful assistants for automating spreadsheet tasks. Users of Copilot and similar agents find these tools useful for reducing the effort of searching online, but they do not necessarily improve task completion time or success rates~\cite{vaithilingam2022expectation}. A critical reason for this shortcoming is that the benchmarks~\cite{li2024sheetcopilot,chen2024sheetagent,payan2023instructexcel} used to evaluate these agents do not truly reflect real and challenging user demands. The primary limitations could be summarized as follows. %These benchmarks often overlook complex instructions that need additional context, such as previous user attempts and encountered issues for explanation. Additionally, they fail to account for the complexity of spreadsheets with flexibly organized data that do not conform to regular relational tables. %In other words, current benchmarks for spreadsheet manipulation fail to accurately align with the practical queries of real users, presenting three main limitations.

\begin{figure}
  \centering
  \includegraphics[width=\textwidth]{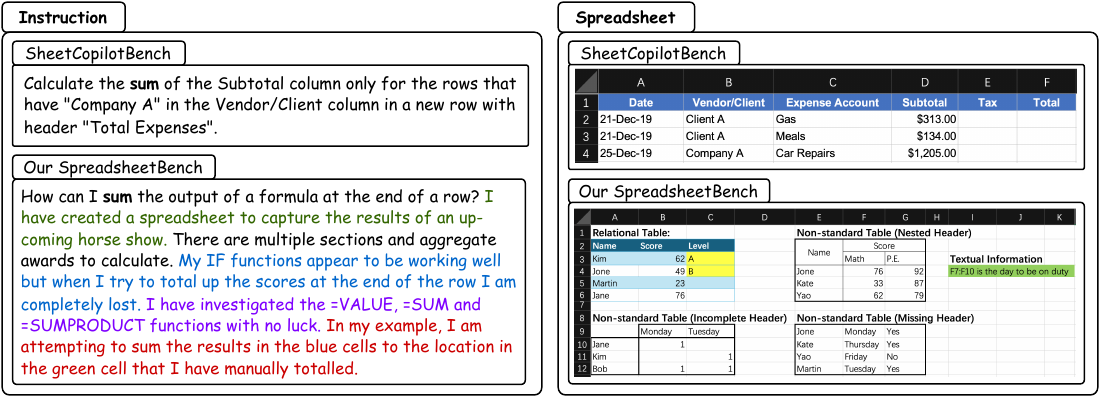}
  \caption{Comparison of previous SheetCopilotBench and our benchmark. The instructions in our benchmark are more complex, reflecting \textcolor[HTML]{336600}{real user demands}, \textcolor[HTML]{0066CC}{issues encountered}, \textcolor[HTML]{7F00FF}{previous user attempts}, and \textcolor[HTML]{CC0000}{an output example}. The spreadsheets in our benchmark organize data more flexibly.
  The figure provides a summary of the various data organization types used in our spreadsheets, with three non-standard relational tables featuring nested header, incomplete header and missing header, in addition to cells containing pure textual information and non-textual elements like colors.}
 \label{fig:data_example} 
\end{figure}

% Current Benchmark

First, these benchmarks either use the self-instruct technique~\cite{self-instruct} to synthesize queries based on a limited set of real-world spreadsheet manipulation demands~\cite{li2024sheetcopilot, chen2024sheetagent}, or they rely on crowdworkers to manually create instructions for given spreadsheets based on a few manipulation examples~\cite{payan2023instructexcel}. As illustrated in the left part of Figure~\ref{fig:data_example}, real user questions collected from online forums (e.g., excelforum.com) differ significantly from synthetic queries, presenting complex demands that often require additional context, such as the users' previous attempts and encountered issues.

Second, the spreadsheets used in these benchmarks are overly simplified~\cite{li2024sheetcopilot, chen2024sheetagent}, typically containing only one regular relational table, similar to the tables used in TableQA tasks~\cite{pasupat2015wtq, nan2022fetaqa} and the tables in databases of Text2SQL tasks~\cite{zhong2017wikisql, yu2018spider}. While both spreadsheets and relational tables store tabular data in a two-dimensional structure, there are significant differences that make spreadsheets more challenging to handle. As shown in  the right part of Figure~\ref{fig:data_example}, spreadsheets offer a flexible way of organizing data without distinct table boundaries or structures, permitting the inclusion of multiple tables in non-standard formats—such as nested, incomplete, or missing headers—and the presence of free-form text within a single cell. Additionally, spreadsheets support various non-textual elements (e.g., color, bold) compared to the relational tables in databases~\cite{zhong2017wikisql} or the CSV/JSON format tables in TableQA tasks~\cite{pasupat2015wtq, katsis2022ait-qa}. These distinct features are not fully utilized in current benchmarks.

Third, these benchmarks for LLMs typically involve a single test for each instruction, where the predicted answer, derived from the solution generated by the LLMs applied to the given spreadsheet, is compared to the ground truth answer~\cite{li2024sheetcopilot, chen2024sheetagent}. This evaluation, however, may result in false positive solutions that are tailored to the specific spreadsheet and do not generalize to other spreadsheets with varying values. A robust solution from LLMs should be able to manage multiple spreadsheets, even corner cases, as inputs, apply the solution, and consistently yield correct answers.

% Our Benchmark
To address the identified need, we introduce a challenging benchmark called \bench. 
Our benchmark features complex user instructions alongside spreadsheets that are flexibly organized by real-world users. These resources are gathered exclusively from four online forums for addressing Excel problems, which are highly ranked by several reputable websites including general Google Search, Feedspot.com---known for its forum search functionality---and Deskbright.com, a site focused on online education, particularly in Microsoft Excel. The instructions often require detailed contextual information for clarification, including descriptions of attempted solutions, incorrect answers received, and the  issues encountered. The spreadsheets exhibit diverse formats, potentially featuring multiple tables within a single spreadsheet, non-standard tables that extend beyond conventional relational structures, cells with only textual content, and other non-textual elements. 

We also propose an Online Judge (OJ)-style metric for more reliable benchmarking, adapted from coding skill assessments~\cite{wasik2018oj}. This metric is well-suited for spreadsheet manipulation tasks where LLMs are tasked with generating code solutions. Within our benchmark, each instruction is associated with multiple input-output spreadsheets serving as test cases, sharing a similar structure but differing in data. A code solution must pass all these test cases to ensure evaluation reliability and accuracy. 

Conclusively, we have developed \bench comprising 912 instructions and 2,729 test cases, with an average of three test cases per instruction. These instructions cover 10 primary manipulation categories and involve spreadsheets with tables extending beyond 100 columns and 20,000 rows. Moreover, 35.7\% of the spreadsheets contain multiple tables, and 42.7\% feature non-standard relational tables. This benchmark is notable for \textbf{its real-world instructions, diverse spreadsheet formats, and a comprehensive testing strategy}.

% Baseline Evaluation
% We conduct a comprehensive evaluation of various types of models, including specifically fine-tuned small models such as TaPas~\cite{herzig2020tapas}, prompt-based LLMs for TableQA like Binder~\cite{Binder}, open-sourced LLMs for coding tasks such as CodeLlama~\cite{roziere2023codellama}, general open-source LLMs (e.g., Llama 3~\cite{llama3modelcard}), the most advanced closed-source models like GPT-4~\cite{achiam2023gpt-4}, and spreadsheet-specific LLMs, including SheetCopilot~\cite{li2024sheetcopilot}. 
We conduct a thorough evaluation of various types of models, including commonly studied TableQA models, open-source LLMs for general tasks and coding tasks, the most advanced close-source models, and spreadsheet-specific LLMs.
The performance of these models vary widely, with scores ranging from 0.05\% to 23.65\% according to our proposed OJ-style evaluation metric. Notably, some methods score as low as 0\%, highlighting the benchmark's difficulty. The results also suggest the importance of improving coding abilities within LLMs for spreadsheet manipulation, and indicate that multi-round prompting, allowing LLMs to read spreadsheet data and error feedback from a code compiler, can increase the probability of correct responses.

%% file: latex/2.benchmark.tex
In this section, we begin by outlining the task formulation, followed by an introduction to the benchmark construction pipeline and the presentation of resulting benchmark statistics. Finally, we introduce the proposed OJ-style evaluation metric.

\subsection{Task Formulation}
\label{sec:problem}

We denote the dataset of a benchmark as $\mathcal{D}=\{(q_i, c_i, a_i)\}$, where $q_i$ represents an instruction, $c_i$ denotes a spreadsheet, and $a_i$ is the corresponding answer of $q_i$ on $c_i$. The objective of spreadsheet manipulation is to enable an LLM to generate a code-based solution $s_i$, apply it to $q_i$ to obtain a predicted answer $\hat{a}_i$, which can then be compared with the ground truth answer $a_i$.

In this benchmark, an instruction $q_i$ involves the modification of either several cells or the entire sheet in a spreadsheet file, which we refer to as \textit{cell-level} and \textit{sheet-level} manipulation, respectively.

The spreadsheet $c_i$ stores tabular data in a two-dimensional format, offering a flexible way for data organization without a distinct table boundary or structure. A single sheet in a spreadsheet file can contain multiple tables in diverse formats, free-form textual information, and non-textual elements. 

An answer $a_i$ refers to the modified spreadsheet resulting from the application of a solution, which is expected to be a piece of code generated by LLMs. We generate code as solutions, as the goal is to manipulate spreadsheets rather than query information from them. Note that $\mathcal{D}$ does not include solutions because we allow for various solutions by LLMs, with our focus solely on the correctness of final answers after applying the solutions.

\subsection{Benchmark Construction}
\label{sec:benchconstruction}
\begin{figure}
  \centering
  \includegraphics[width=\textwidth]{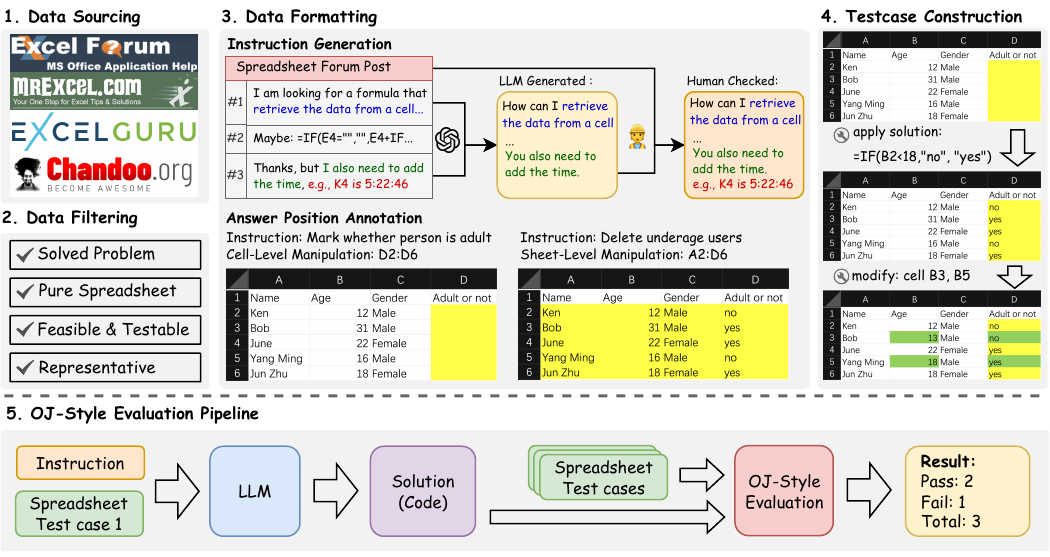}
  \caption{The benchmark construction pipeline and OJ-style evaluation of our benchmark.} 
 \label{fig:pipeline} 
\end{figure}

The construction of the benchmark follows a four-stage pipeline: 1) data sourcing, 2) data filtering, 3) data formatting, and 4) test case construction, as illustrated in Figure~\ref{fig:pipeline}.
Subsequently, we detail the perturbation techniques used to prevent data leakage.

% \paragraph{1. Data Sourcing}
\textbf{1. Data Sourcing.}
Our objective is to collect high-quality and representative spreadsheet manipulation questions from real-world sources. As depicted in the initial section of Figure~\ref{fig:pipeline}, we utilize general Google Search, the specialized forum search engine Feedspot.com, and the Microsoft Excel-focused online learning platform Deskbright.com to identify four popular and regularly updated forums and blog sites—ExcelForum, Chandoo, MrExcel, and ExcelGuru—as our data sources (detailed information available in Appendix~\ref{app:forum}).
We specifically target posts that fall under categories such as \textit{General Questions}, \textit{Formula}, \textit{VBA \& Macros}, \textit{Formatting}, \textit{Pivot Table}, and \textit{Power Query} to ensure relevance to spreadsheet manipulation. %Ultimately, we have gathered \textcolor{Red}{xxx} posts and \textcolor{Red}{xxx} blogs for further filtering.

% \paragraph{2. Data Filtering}
\textbf{2. Data Filtering.}
Since the original posts in forums are diverse with various questions, as shown in the second part of Figure~\ref{fig:pipeline}, we select the question by four main criteria: 1) is solved, 2) solely pertains to spreadsheet manipulation, 3) is feasible and testable, and 4) is representative.

Initially, we directly identify questions  tagged as ``solved'' from all posts. For posts without explicit tags, we employ GPT-4 to determine their solved status, as patterns such as thank-you notes are often observed at the end of a post indicating the question is solved.

Next, we focus on extracting questions strictly related to spreadsheet manipulation. Most popular spreadsheet forums primarily discuss Excel-related issues involving software functions like input boxes, buttons, and forms. Unlike InstructExcel~\cite{payan2023instructexcel}, our goal is to establish a software-independent benchmark centered solely on spreadsheet manipulation. Consequently, we discard all irrelevant software-specific questions using keywords such as ``input boxes'', ``buttons'', and ``forms'', and engage annotators to check the remaining posts.

Finally, we select questions that are both feasible and representative. To ensure feasibility, aided by annotators, we exclude posts without spreadsheet attachments, those with excessive replies, or ambiguous presentations. To identify representative posts, we filter those with high view counts and ask the annotators to retain both simple and difficult questions.

% \paragraph{3. Data Formatting}
\textbf{3. Data Formatting.}
According to the task definition outlined in Section~\ref{sec:problem}, it is necessary to formulate the instruction, the spreadsheet, and the answer in a manner that facilitates convenient and accurate automatic evaluation.

\textit{Instruction Generation.} The original posts are often cluttered, with the description of the user question dispersed across multiple replies. Therefore, we need to extract and condense these into a self-contained instruction from all replies. Initially, we utilize GPT-4 to recreate a coherent instruction from the original post. Annotators then verify this instruction to resolve any issues arising from possible incomplete context extraction from the post.

\textit{Answer Position Annotation.} Regarding solution derivation, forum users typically discuss solutions in formulas or VBA code. Annotators are responsible for  identifying the solution from the replies and applying it to the corresponding spreadsheet to derive the answers. However, discrepancies in the placement of answers by annotators compared to those predicted by LLMs can lead to mistaken evaluations. To mitigate this issue, we allow annotators to incorporate specific constraints in the instructions that annotate the answer position. As described in Section~\ref{sec:problem}, answer positioning should consider both cell-level and sheet-level manipulation. For cell-level manipulations, annotators should specify exact cell positions, such as ``D2:D6''. For sheet-level manipulations, the entire range of the spreadsheet, such as ``A2:D6'', needs to be included in the instruction.
\begin{figure}
  \centering
  \includegraphics[width=\textwidth]{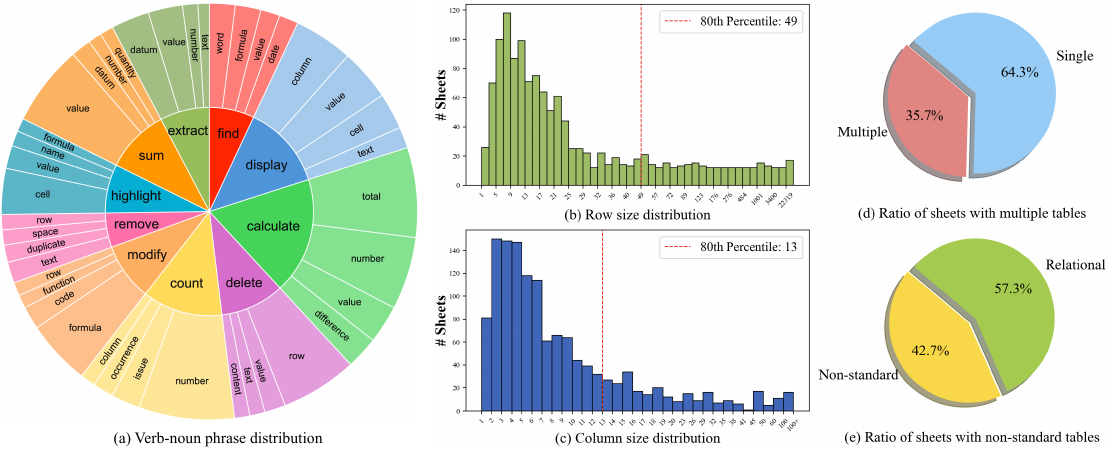}
  \caption{Key statistics of \bench.
% : 
% (a) The diversity of verb-noun phrases in our instructions by listing the top ten most frequent root verbs (inner circle) and their corresponding noun objects (outer circle).
% (b) The distribution of the row and column sizes of our spreadsheet files.
% (c) The ratio of sheets with multiple tables and non-standard relational tables in our spreadsheet files.
} 
 \label{fig:datastat} 
\end{figure}

% \paragraph{4. Test Case Construction}
\textbf{4. Test Case Construction.}
According to observations on online forums, it is evident that some solutions provided by forum users may be applicable only to the specific example spreadsheet provided, potentially failing when applied to other scenarios with slight variations in the spreadsheet data.
To enhance the reliability of user-provided solutions, it is essential to create multiple spreadsheets and develop multiple test cases for each instruction. This preparation is essential for the subsequent OJ-style evaluation metric.

Specifically, for instructions that solely offer a solution without derived answers from forum posts, we derive the answer for the original provided spreadsheet by manually manipulating the input spreadsheet using the given solution presented by formulas or VBA code. The original spreadsheet, along with the derived answer, forms an input-output test case for the current instruction.

Subsequently, annotators are requested to modify the data in the original spreadsheet twice to generate two distinct spreadsheets. By applying the same solution to each of these modified spreadsheets, we can obtain two corresponding answers. 
It's important to note that annotators do not make casual alterations to the spreadsheet data; instead, they are instructed to concentrate on real-world corner cases (e.g., cells that should contain zeros but remain empty).
Annotators are also instructed to make valid modifications consistent with the question's requirements.

Ultimately, for each instruction suitable for multiple test cases, we derive three test cases from the original single spreadsheet-answer data, as illustrated in the fourth part of Figure~\ref{fig:pipeline}. Then $\mathcal{D}=\{(q_i,c_i,a_i)\}$ is changed to $\mathcal{D}=\{(q_i,\{(c_{ij},a_{ij})\})\}$.

% \paragraph{Against Data Leakage}
\textbf{Against Data Leakage.}
Datasets initially obtained from online forums may be susceptible to data leakage issues, given that many LLMs are pre-trained using a vast corpus of web text~\cite{elangovan2021data-leak,lai2023ds1000}.

The above mentioned strategies, along with some new approaches, can introduce perturbations to the collected data to mitigate the data leakage problem:

\textit{Instruction Generation}:
The aforementioned instruction generation strategy, carried out by GPT-4 and human annotators, involves revising the original questions in the posts, thereby preventing LLMs from memorizing the original questions.

\textit{Spreadsheet Modification}:
The test case construction strategy outlined above involves modifying the original provided spreadsheets, preventing LLMs from memorizing the original spreadsheets.

\textit{Answer Position Changing}:
We also alter the position of the tabular data in the original spreadsheets and the corresponding answer in the resulting spreadsheets. By doing so, the originally provided solution from the posts cannot be directly used to derive answers with changed positions, thus preventing LLMs from memorizing the provided solution from the posts.

% 根据审稿人意见，详细介绍如何保证软件独立性
\textbf{Software Independence.}
Spreadsheet manipulation is not inherently software-independent, as some functions may not be available (or may change name) changing software.
We employ three methods to ensure the software independence of our benchmark as much as possible.
(1) \textit{Software-independent instruction.}
We only select questions that focus on the manipulation of the value of cells in the spreadsheet and try to avoid questions involving special components or advanced operations that only exist in specific software.
(2) \textit{Software-independent spreadsheet files.}
All the spreadsheet files in our dataset are in the XLSX format, which is widely supported across multiple spreadsheet applications, such as Microsoft Excel, Google Sheets, LibreOffice Calc, and WPS Office.
Though some spreadsheet software may not provide full support for the XLSX format, we have made an effort to exclude any content that may be incompatible with non-Excel software.
(3) \textit{Software-independent Inference.}
Instead of utilizing software-dependent tools (e.g., TypeScript and the Excel API) to interact with spreadsheet files\cite{payan2023instructexcel}, we allow models to manipulate spreadsheet files in diverse ways, such as Python code, because we only compare whether the results after manipulations are correct.

% 根据审稿人意见，将标注团队的信息添加到正文中
\textbf{Annotation Team.}
The construction of our benchmark requires high-quality human annotations.
We employ a team of 20 individuals who specialize in Excel and have extensive experience in annotation.
All team members hold a bachelor’s degree and are compensated in line with market rates.
For data annotation, we allocate a predetermined quantity of data to annotators on a daily basis, and the allocation is modified in accordance with local regulations regarding work hours.
For data validation, we hire two experienced validators with bachelor's degrees for the first round of quality checks.
Moreover, two authors who hold master's degrees perform a secondary round of quality assessments to guarantee the high quality of data.

\subsection{Benchmark Statistics}
\label{sec:statistic}

% \paragraph{Data Statistics}
\textbf{Data Statistics.}
We have developed \bench comprising 912 instructions and 2,729 test cases, with an average of three test cases per instruction. 
We analyze it and derive several observations, as depicted in  Figure~\ref{fig:datastat}. In Figure~\ref{fig:datastat}(a), the instructions in our benchmark cover a broad spectrum of spreadsheet manipulation types, including find, extract, sum, highlight, remove, modify, count, delete, calculate, and display. The outer circle illustrates the objects being manipulated, showcasing the diversity of the benchmark.

Figures~\ref{fig:datastat}(b) and (c) display the distributions of row and column sizes in our spreadsheet files, revealing long-tailed distributions. Specifically, 80\% of spreadsheets have a row size spanning from 1 to 49, and a column size spanning from 1 to 13. It is noteworthy that the row or column size in the long tail part is substantial, indicating the challenging nature of our benchmark. % due to the inclusion of a significant number of spreadsheets with a large number of rows and columns.

Figure~\ref{fig:datastat}(d) demonstrates that more than one-third of spreadsheets contain multiple tables in one sheet. Figure~\ref{fig:datastat}(e) reveals that nearly half of the spreadsheets contain non-standard relational tables with nested, incomplete or missing headers.
These observations indicate that the \bench significantly increases the complexity of understanding and manipulating the spreadsheets.

\input{table/data_statistics}
% \paragraph{Comparison with Previous Benchmarks}
\textbf{Comparison with Previous Benchmarks.}
Table~\ref{tab:data_statistics} compares \bench with previous spreadsheet manipulation benchmarks. 
In addition, numerous benchmarks exist for table-related tasks. These include table question answering (TableQA)\cite{pasupat2015wtq,zhu2021tat-qa, chen2021ottqa, cheng2022hitab}, Table-to-text\cite{zhao2023loft, zhao2023investigating}, and table data analysis~\cite{he2024text2analysis, li2024tapilot,hu2024infiagent}. We have not included benchmarks for these table-related tasks, as their instructions are specifically tailored for QA and use CSV/JSON files as input, which significantly diverges from tasks that involve spreadsheet manipulation.

\bench is sourced exclusively from real-world data and exhibits a higher average word count per instruction.
%, presenting a wider range of diverse and realistic demands. 
We have collected a larger number of spreadsheet files, including both single-sheet and multiple-sheet formats in each file. Our spreadsheet files contain multiple sheets with non-standard relational tables and multiple tables within a single sheet. 
%posing significant challenges for spreadsheet agents to handle.
Real-world questions often involve additional explanations within the spreadsheet, a characteristic not present in previous benchmarks. Furthermore, we employ OJ-style evaluation metrics with three test cases per instruction.
%offering more accuracy and granularity compared to pure exact match (EM) and similarity metrics. 

The benchmarks SheetCopilotBench~\cite{li2024sheetcopilot} and SheetRM~\cite{payan2023instructexcel} generate instructions using LLMs' self-instruct techniques, which are quite different from users' actual queries. Furthermore, the manipulated spreadsheets in these benchmarks are relatively simple, seldom featuring multiple sheets and tables, and non-standard relational tables. On the other hand, InstructExcel~\cite{payan2023instructexcel} offers a more comprehensive benchmark, particularly with complex spreadsheet files that cover various features. Even though the instructions in InstructExcel are created by annotators, they remain relatively simple. The average word count per instruction is only 9.8, compared to 85.7 in our benchmark. Additionally, InstructExcel does not include additional explanations or multiple test cases for the instructions, making it less complex and potentially less reliable than our benchmark.

\subsection{Evaluation Metrics}
\label{sub_sec:evaluation_metrics}
Exact match is a commonly employed metric for spreadsheet manipulation tasks~\cite{li2024sheetcopilot, chen2024sheetagent}, but it may only work for the current spreadsheets and may not be robust to variations in the data within the spreadsheet.
To address this, we adopt an approach similar to the online judge system commonly used for assessing coding capability~\cite{wasik2018oj}.
An Online Judge (OJ) style evaluation involves inputting multiple test cases into a code script and comparing each test case's output to the ground truth result.
This method is more comprehensive and reliable than evaluating the code script just once.
In our benchmark context, each test case is a spreadsheet file with a similar table structure but different cell values, as mentioned in Section~\ref{sec:benchconstruction}.
This allows us to evaluate a code solution more thoroughly and determine its effectiveness in handling corner cases.
% We prepare multiple spreadsheet files as test cases for each instruction, as mentioned in Section~\ref{sec:benchconstruction}, and employ a similar OJ-style metric to ensure the reliability of the evaluation process.
Figure~\ref{fig:pipeline} illustrates this process, where an LLM is expected to generate a solution given an instruction and a spreadsheet test case.
Subsequently, the solution is applied to multiple spreadsheet test cases for OJ-style evaluation.
In contrast to some benchmarks that perturb data in tables to create new data samples and perform one model inference for each data sample~\cite{zhou2024freb-tqa}, we only require one model inference to generate a solution for all test cases of each instruction. 

We establish two distinct scoring criteria to calculate the final score. The \textit{soft restriction} adheres to the scoring principles of the OJ system from the IOI, granting partial credit when a solution only passes some test cases~\cite{cormack2006oj_detail}. The calculation is as follows:

\begin{equation}
\label{eq:soft}
    S_{soft}=\frac{1}{|\mathcal{D}|}\sum_{i=1}^{|\mathcal{D}|}\left(\frac{1}{|T_i|}\sum_{j=1}^{|T_i|}\mathds{1}_{r_{ij} = ACC}\right).
\end{equation}

The \textit{hard restriction} follows the ICPC scoring rules of the OJ system, where no partial credit is awarded~\cite{cormack2006oj_detail}. The score is determined as follows:

\begin{equation}
\label{eq:hard}
    S_{hard}=\frac{1}{|\mathcal{D}|}\sum_{i=1}^{|\mathcal{D}|}\mathds{1}_{r_{ij} = ACC, \forall j = 1, 2, \ldots, |T_i|}.
\end{equation}

In the above equations, $T_i$ represents the test cases (i.e., $\{(c_{ij}, a_{ij})\}$) of the $i$-th instruction, and  $r_{ij}$ indicates the result of the model's solution applied on the $j$-th test case, respectively. Specifically, $r_{ij}$ is marked as Accept (ACC) if the applied solution's answer on the $j$-th test case (i.e., the resulting spreadsheet) match the ground truth answer. The soft restriction metric does not penalize models for failing to provide a flawless solution that addresses every common and corner cases, whereas the hard restriction metric encourages models to strive for the most perfect solution.

%% file: table/data_statistics.tex
\begin{table}
  \caption{Comparison between \bench and other spreadsheet manipulation benchmarks across four aspects: data source, instructions, spreadsheets, and evaluation metrics.}
  \label{tab:data_statistics}
  \centering
  \resizebox{\linewidth}{!}{
  \begin{tabular}{l|ccc|c}
    \toprule
    Benchmark & SheetCopilotBench~\cite{li2024sheetcopilot} & InstructExcel~\cite{payan2023instructexcel} & SheetRM~\cite{chen2024sheetagent} & Ours \\
    \midrule
    Data Source & Self-Instruct & Manual Annotation & Self-Instruct & Forum \& Blog \\
    \midrule
    Instructions & 221 & 4850 & 201 & 912 \\
    Ave. Instruction Words & 27.9 & 9.8 & - & 85.7 \\
    \midrule
    Spreadsheet Files & 29 & 940 & 25 & 2729 \\
    \quad Single Sheet & 26 & 572 & - & 2019 \\
    \quad Multiple Sheet & 3 & 368 & - & 710 \\
    Sheets & 32 & 1694 & 83 & 3917 \\
    \quad Non-standard Tables & \ding{56} & \textcolor{Red}{\ding{51}} & \ding{56} & \textcolor{Red}{\ding{51}} \\
    \quad Multiple Tables & \ding{56} & \textcolor{Red}{\ding{51}} & \ding{56} & \textcolor{Red}{\ding{51}} \\
    \quad Additional Info. & \ding{56} & \ding{56} & \ding{56} & \textcolor{Red}{\ding{51}} \\
    \midrule
    Evaluation & Exact Match (EM) & EM \& Similarity & Sub-task EM & OJ-style EM \\
    Ave. Test Cases & 1 & 1 & 1 & 3 \\
    \bottomrule
  \end{tabular}}
\end{table}

%% file: latex/3.experiments.tex
\label{sec:exp_setup}
\textbf{Baselines.}
We evaluate LLMs across five categories: (1) \textit{TableQA models}, including fine-tuned TaPEx (based on BERT)~\cite{liu2022tapex}, TaPas (based on BART)~\cite{herzig2020tapas}, and prompt-based  Binder (GPT-3.5)~\cite{Binder}; (2) \textit{Open-source code models}, including CodeQwen (7B)~\cite{qwen} and DeepseekCoder (33B)~\cite{deepseek-coder}; (3) \textit{Open-source general models}, including Mixtral 8x7B~\cite{jiang2024mixtral} and Llama 3 (70B)~\cite{llama3modelcard}; (4) \textit{Close-source models}, including GPT-3.5\footnote{\url{https://platform.openai.com/overview}} and GPT-4o~\cite{achiam2023gpt-4}; (5) \textit{Spreadsheet-specific methods or products}, including SheetCopilot~\cite{li2024sheetcopilot} and Copilot in Excel\footnote{\url{https://www.microsoft.com/microsoft-copilot/}}.

% 审稿人要求加上模型是如何和电子表格交互的，对不了解此领域的人比较友好
\textbf{Evaluation Interface.}
For TableQA models, we ask the model to generate text results based on the content of the spreadsheet files.
For Spreadsheet-specific methods or products, we manually evaluate these methods within their products.
We only sample five percent of the data samples from our benchmark due to the costly and time-consuming process of evaluating the entire benchmark.
For other models, we use Python code to modify and interact with the spreadsheet files.

% \paragraph{Inference Setting}
\textbf{Inference Setting.}
We evaluate LLMs under two distinct settings:
1. \textit{Single Round}: In this mode, we present the model with the initial few rows of spreadsheet files within the prompt, allowing for only one inference. Since TableQA models can only produce textual answers, we conduct separate inferences for each test case and compare the resulting textual answers to the ground truth. For non-TableQA models, we instruct them to generate a code-based solution for all test cases of a given instruction.
2. \textit{Multi-Round}: Building on the single-round prompt setting, we incorporate additional prompt that utilizes the ReAct technique~\cite{yao2022react} and code execution feedback~\cite{zheng2024opencodeinterpreter} to enhance the accuracy of code solutions produced by LLMs over multi-round conversation. Specifically, we offer LLMs two choices: to generate code for retrieving spreadsheet file contents or to provide the ultimate solution. For the first choice, we supply code execution outcome to the subsequent round. In the second, the conversation concludes once the desired spreadsheet file is produced. In both instances, we furnish error feedback if the code fails to execute, enabling the model to refine its code in subsequent iterations. We impose a limit of five rounds for the multi-round setting.

% 按审稿人要求，添加human performance内容
\textbf{Human Performance.}
We hire four annotators who are experts in Excel to obtain the human performance of our benchmark.
To minimize costs, we use a subset of fifty instructions and three test cases for each instruction to conduct our human evaluation.
Annotators are asked to provide formula solutions or VBA code to finish each instruction instead of filling in the answer in the target position directly.
The process of human evaluation is completed within one day.

% (3) \textit{Code Interpreter}\footnote{\url{https://platform.openai.com/docs/assistants/tools/code-interpreter}} is a tool provided by OpenAI, allowing AI assistants to to write and execute Python code within a secure environment.
% It is designed to handle files containing various types of data and formatting.
\input{table/main_result}

% \subsection{Results and Analysis}

% \paragraph{Overall performance}
\textbf{Overall Performance.}
The results shown in Table~\ref{tab:main_result} indicate that current LLMs and spreadsheet agents are inadequate in managing complex spreadsheet manipulation tasks as required by real-world scenarios. Even the most advanced spreadsheet agent, Copilot in Excel, only achieves an accuracy of roughly 20\%. GPT-4o, the SOTA LLM, scores around 17\% in accuracy, aligning with Copilot in Excel's performance. Open-source LLMs significantly underperform compared to the SOTA model, likely due to their limited comprehension and coding proficiency.
Binder shows the poorest results, while TaPEx and TaPas, designed specifically for TableQA, score zero across all metrics (omitted from the table). This underscores the distinction in difficulty between TableQA and spreadsheet manipulation.
Overall, there is a substantial gap between existing LLMs or products and human performance produced by Excel experts, emphasizing the critical need for advancement in LLMs tailored for spreadsheet manipulation. 

\input{table/result_analysis}

Alongside the performance evaluation of each model, our findings include the following insights:

(1) \textit{Code Model vs. General Model.}
Given that spreadsheet manipulation solutions primarily rely on coding, LLMs tailored for coding, such as DeepseekCoder, exhibit superior performance compared to open-source general models like Llama-3 (70B). This highlights the need to enhance coding capacities within LLMs for effective spreadsheet manipulation.

(2) \textit{Single Round vs. Multiple Round.}
The majority of LLMs experience significant improvement in their capabilities through multi-round conversation, with some models achieving nearly a tenfold increase.
However, GPT-4o’s performance slightly declines in a multi-round setting. This is due to GPT-4o’s adherence to instructions, which leads to duplicated content when retrieving the spreadsheet, as it includes the initial spreadsheet rows already provided in the prompt. Other LLMs, unable to follow the retrieval instruction, rely on the given initial rows. We provide the performance of GPT-4o without the initial spreadsheet rows in Appendix~\ref{app:detail_exp}.

(3) \textit{Soft Restriction vs. Hard Restriction.}
The shift from soft to hard restrictions leads to a modest decrease in LLM performance, suggesting that the solutions produced by these models may not remain effective when the spreadsheet content is altered. Given that real-world spreadsheet applications often involve frequent content changes, it is crucial to develop solutions that are robust to such modifications. The introduced OJ-style evaluation metrics are well-suited to assess such robustness.

% \paragraph{Analysis}
\textbf{Analysis.}
We also conduct analysis on the factors that influence the performance of LLMs on the spreadsheet manipulation task.

(1) \textit{Great spreadsheet complexity leads to diminished performance.} 
Our benchmark is organized into four categories based on row and column size, the presence of multiple tables, and the inclusion of non-standard tables, resulting in eight distinct subsets, as detailed in Table~\ref{tab:subsetaccuracy}. Our analysis reveals that model performance notably declines when dealing with spreadsheets that have extensive rows and columns, multiple tables, and non-standard structures.
% Given the significant presence of these complex spreadsheets in real scenarios, it is essential to empower LLMs with the capability to effectively manipulate them.

(2) \textit{Increase row size, no significant performance gain.}
We assess the effect of row size in the prompt under a single round setting. Figure~\ref{fig:impact_factor}(a) shows that expanding input rows from 5 to 10 doesn’t notably enhance performance. This could be due to the excessive context length when processing additional rows.

(3) \textit{GPT4o does not benefit from the multi-round setting.}
As shown in Figure~\ref{fig:impact_factor}(b), the performance of GPT-3.5 improves as the round number increases from 2 to 4.
However, GPT-4o, the current most advanced LLM, does not benefit from the multi-round setting.
Our analysis shows that GPT-4o's initial solutions have a substantially higher rate of executability and accuracy than other models. Consequently, the additional step of repeatedly retrieving spreadsheet content and executing code does not markedly improve GPT-4o's performance.

\textbf{Case Study.}
We provide two cases (Figure~\ref{fig:case_study_1-1} and Figure~\ref{fig:case_study_2-1}) and conduct qualitative analysis of the LLM's performance on different characteristics of the spreadsheet.
To get the code solution of these two cases, we utilize the multi-round setting that is discussed in the inference setting section.

Example 1 shows an instruction that manipulates a non-standard relational table with formatting in spreadsheet.
Both GPT-3.5 and GPT-4 understand the question and read the spreadsheet properly.
However, due to the lack of coding ability, GPT-3.5 uses the wrong element \texttt{``correct\_results[cell.column]''} (marked as red), as shown in Figure~\ref{fig:case_study_1-2}.
The correct one should be \texttt{``correct\_results[cell.column - 1].value''}.
In Figure~\ref{fig:case_study_1-3}, GPT-4o can generate the correct Python code, highlighting the importance of coding ability for the accuracy of spreadsheet manipulation.

Example 2 shows an instruction that manipulates two tables in one sheet with formatting.
The two GPT-4 inference results both show issues with understanding the structure of the spreadsheet.
For the first result of GPT-4o in Figure~\ref{fig:case_study_2-2}, misalignment occures when reading the tables. ``5A1'' belong to the column names ``Green'' but is highlighted in yellow.
``5A3'' belong to the column names ``Amber'' but is highlighted in red.
For the second result of GPT-4o in Figure~\ref{fig:case_study_2-3}, the model did not read the whole lookup table in column E to G.
The rows after the tenth row have been ignored.
Thus, ``1A4'', ``D2'', and ``1A1'' are not highlighted.

These two examples show that proficient coding skills and a solid understanding of various spreadsheet structures are essential for the spreadsheet manipulation task.

%% file: table/main_result.tex
\begin{table}
  \caption{Performance of representative models on \bench (\%).}
  \label{tab:main_result}
  \centering
  \resizebox{\linewidth}{!}{
  \begin{tabular}{lcccccc}
    \toprule
    \multirow{2}{*}[-3pt]{Model} & \multicolumn{3}{c}{Soft Restriction ($\uparrow$)} & \multicolumn{3}{c}{Hard Restriction ($\uparrow$)}\\
    \cmidrule(r){2-4}\cmidrule(r){5-7}
    & Cell-Level & Sheet-Level & Overall & Cell-Level & Sheet-Level & Overall \\
    \midrule
    % TaPEx  &      \\
    % TaPas  &      \\
    % \midrule
    Binder (GPT-3.5) & 1.58 & 0.05 & 1.17 & 0.00 & 0.00 & 0.00   \\
    \midrule
    CodeQwen (7B)  & 0.36 & 0.76 & 0.51 & 0.36 & 0.29 & 0.33 \\
    \quad w / Multi-Round & 1.49 & 7.14 & 3.66 & 0.89 & 6.29 & 2.97 \\
    DeepseekCoder (33B)  &  0.59 & 5.81 & 2.60 & 0.36 & 5.14 & 2.20 \\
    \quad w / Multi-Round  &  3.15 & 8.76 & 5.31 & 1.96 & 6.86 & 3.85 \\
    \midrule
    Mixtral-8x7B & 2.97 & 3.33 & 3.11 & 2.32 & 2.57 & 2.42 \\
    \quad w / Multi-Round & 3.39 & 4.67 & 3.88 & 2.32 & 3.71 & 2.85 \\
    Llama-3 (70B)  & 0.18 & 3.14 & 1.32 & 0.00 & 2.86 & 1.10 \\
    \quad w / Multi-Round &  1.13 & 7.90 & 3.74 & 0.71 & 7.14 & 3.18 \\
    \midrule
    GPT-3.5 & 1.31 & 3.99 & 2.34 & 0.71 & 3.13 & 1.64 \\
    \quad w / Multi-Round & 3.33 & 13.11 & 7.09 & 2.50 & 9.97 & 5.37 \\
    GPT-4o & 15.03 & \textbf{23.65} & 18.35 & \textbf{11.94} & \textbf{19.94} & \textbf{15.02} \\
    \quad w / Multi-Round & 13.49 & 22.51 & 16.96 & 10.52 & 17.66 & 13.27 \\
    \midrule
    SheetCopilot (GPT-4)* & 16.67 & 10.00 & 14.00 & - & - & - \\
    Copilot in Excel* & \textbf{23.33} & 15.00 & \textbf{20.00} & - & - & - \\
    \midrule
    \rowcolor{gray!20}
    Human Performance & 75.56 & 65.00 & 71.33 & 66.67 & 55.00 & 62.00 \\
    \bottomrule
  \end{tabular}}
\end{table}

%% file: table/result_analysis.tex
\begin{figure}[t]
    \centering
        \begin{minipage}{0.37\linewidth}
	\centering
        \small
        \resizebox{0.95\linewidth}{!}{
        \begin{tabular}{lcc}
            \toprule
            Task Subset & \% of Total & Accuracy \\ 
            \midrule
            Rows ($\leq 50$) & 75.19 & \textbf{20.63} \\ 
            Rows ($> 50$) & 24.81 & 11.50 \\
            \midrule
            Columns ($\leq 10$) & 65.53 & \textbf{22.50} \\
            Columns ($> 10$) & 34.47 & 10.51 \\
            \midrule
            Single Tab. & 62.90 & \textbf{21.12} \\
            Multiple Tab. & 37.10 & 13.71 \\
            \midrule
            Relational Tab. & 55.54 & \textbf{19.50} \\
            Non-standard Tab. & 44.46 & 16.95 \\
            \bottomrule
    \end{tabular}}
        \captionof{table}{Overall soft restriction of GPT-4o on different subsets (\%).}
        \label{tab:subsetaccuracy}
	\end{minipage}
	%\qquad
        % 图片
	\hfill
	\begin{minipage}{0.6\linewidth}
		\centering
		\vspace{-0.16cm}
		\setlength{\abovecaptionskip}{0.73cm}
		\includegraphics[width=\linewidth]{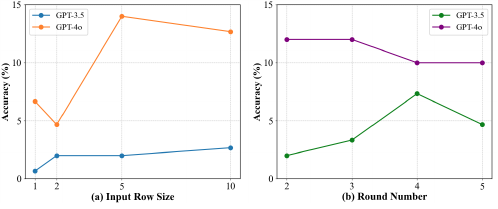}
            \vspace{-10mm}
		\caption{Impact of input row size and round number in terms of overall soft restriction.}
		\label{fig:impact_factor}
	\end{minipage}
    \vspace{-5mm}
\end{figure}

%% file: latex/5.conclusion.tex
We create a rigorous benchmark, \bench, for evaluating LLMs' capacity in spreadsheet manipulating. Compared with existing benchmarks, \bench incorporates 912 authentic instructions sourced from online forums, more diverse spreadsheet files with rich formats, multiple tables, irregular tables, and more comprehensive testing with multiple test suites, ensuring better coverage of corner cases. %We believe \bench could advance the spreadsheet manipulation abilities of LLMs.

%% file: latex/6.checklist.tex
%%% BEGIN INSTRUCTIONS %%%
% The checklist follows the references.  Please
% read the checklist guidelines carefully for information on how to answer these
% questions.  For each question, change the default \answerTODO{} to \answerYes{},
% \answerNo{}, or \answerNA{}.  You are strongly encouraged to include a {\bf
% justification to your answer}, either by referencing the appropriate section of
% your paper or providing a brief inline description.  For example:
% \begin{itemize}
%   \item Did you include the license to the code and datasets? \answerYes{See Section~\ref{gen_inst}.}
%   \item Did you include the license to the code and datasets? \answerNo{The code and the data are proprietary.}
%   \item Did you include the license to the code and datasets? \answerNA{}
% \end{itemize}
% Please do not modify the questions and only use the provided macros for your
% answers.  Note that the Checklist section does not count towards the page
% limit.  In your paper, please delete this instructions block and only keep the
% Checklist section heading above along with the questions/answers below.
%%% END INSTRUCTIONS %%%

\begin{enumerate}

\item For all authors...
\begin{enumerate}
  \item Do the main claims made in the abstract and introduction accurately reflect the paper's contributions and scope?
    \answerYes{See Section~\ref{sec:introduction}}
  \item Did you describe the limitations of your work?
    \answerYes{See Appendix~\ref{app:limitation}}
  \item Did you discuss any potential negative societal impacts of your work?
    \answerYes{See Appendix~\ref{app:potential_impact}}
  \item Have you read the ethics review guidelines and ensured that your paper conforms to them?
    \answerYes{See Appendix~\ref{app:ethical}}
\end{enumerate}

\item If you are including theoretical results...
\begin{enumerate}
  \item Did you state the full set of assumptions of all theoretical results?
    \answerNA{}
	\item Did you include complete proofs of all theoretical results?
    \answerNA{}
\end{enumerate}

\item If you ran experiments (e.g. for benchmarks)...
\begin{enumerate}
  \item Did you include the code, data, and instructions needed to reproduce the main experimental results (either in the supplemental material or as a URL)?
    \answerYes{See Section~\ref{sec:exp_setup} and Appendix~\ref{app:detail_exp}}
  \item Did you specify all the training details (e.g., data splits, hyperparameters, how they were chosen)?
    \answerYes{See Appendix~\ref{app:detail_exp}}
	\item Did you report error bars (e.g., with respect to the random seed after running experiments multiple times)?
    \answerNo{Due to the excessive cost of executing API-based models and methods.}
	\item Did you include the total amount of compute and the type of resources used (e.g., type of GPUs, internal cluster, or cloud provider)?
    \answerYes{See Appendix~\ref{app:detail_exp}}
\end{enumerate}

\item If you are using existing assets (e.g., code, data, models) or curating/releasing new assets...
\begin{enumerate}
  \item If your work uses existing assets, did you cite the creators?
    \answerYes{See Seciton~\ref{sec:introduction}}
  \item Did you mention the license of the assets?
    \answerYes{See Appendix~\ref{app:maintenance}}
  \item Did you include any new assets either in the supplemental material or as a URL?
    \answerYes{See Abstract}
  \item Did you discuss whether and how consent was obtained from people whose data you're using/curating?
    \answerYes{See Appendix~\ref{app:detail_data_collection}}
  \item Did you discuss whether the data you are using/curating contains personally identifiable information or offensive content?
    \answerYes{See Appendix~\ref{app:ethical}}
\end{enumerate}

\item If you used crowdsourcing or conducted research with human subjects...
\begin{enumerate}
  \item Did you include the full text of instructions given to participants and screenshots, if applicable?
    \answerYes{See Appendix~\ref{app:detail_data_collection}}
  \item Did you describe any potential participant risks, with links to Institutional Review Board (IRB) approvals, if applicable?
    \answerNA{}
  \item Did you include the estimated hourly wage paid to participants and the total amount spent on participant compensation?
    \answerYes{See Appendix~\ref{app:ethical}}
\end{enumerate}

\end{enumerate}

%% file: latex/7.appendix.tex
\section{Broader Discussion}

\subsection{Limitation}
\label{app:limitation}
The major limitation of \bench lies in data selection and test case construction process.
(1) \textit{Data Selection}:
As part of our data selection process, we eliminate posts that lack an acknowledged response or are difficult to formalize.
Although the proportion of these posts that contain valuable spreadsheet manipulation questions is low, they may also be appropriate for our benchmark under careful human annotation.
(2) \textit{Test Case Construction}:
Our benchmark has a significantly higher number of questions compared to online judge competitions.
Consequently, we did not meticulously devise corner cases for each question, taking into account all possible scenarios and edge cases.
However, we still request annotators to remain vigilant regarding potential corner cases while annotating each question (See Figure~\ref{fig:example_corner} for an example).

\subsection{Potential Impact}
\label{app:potential_impact}
\bench are constructed based on real user queries and spreadsheet files.
Our objective is to improve the proficiency of LLMs in comprehending and manipulating spreadsheets, with the ultimate aim of automating the spreadsheet manipulation process and reduce human workload in the future.
Besides, our benchmark consists of tabular data and corresponding questions, making it suitable for evaluating the anbility of LLMs in understanding and manipulating structured data.
Our benchmark exclude any content that could violate personal privacy, contain explicit material, depict violence, or involve other sensitive subjects during our human annotation process.
Thus, we believe that the probability of our benchmark causing adverse effects on safety, security, discrimination, surveillance, deception, harassment, human rights, bias, and fairness is extremely minimal.

\subsection{Ethical Consideration}
\label{app:ethical}
In this section, we discuss the ethical considerations regarding our data construction process.
(1) \textit{Data Risk Control}:
During our annotation process, we filter out the content in questions or spreadsheet files that is inappropriate for presentation to a general audience, such as contents that depict violence, sensitive subjects, etc.
In addition, two leaders from the annotation team and two authors conduct an additional verification to ensure that our benchmark does not contain any instances of personally identifiable information.
(2) \textit{Annotator Treatment and Consent}:
We recruit crowdsourced annotators for the construction of our benchmark (See \ref{app:data_annotation} for details).
We have formalized employment agreements with all the annotators and remunerated them according to the predetermined wage criteria and working schedule.
All employment agreements adhere to local regulations.
(3) \textit{Copyright}:
Our benchmark is indirectly derived from posts on public Excel forums and blogs.
The majority of the raw data is acquired from the ExcelForum website using Python crawler scripts, while the data from the remaining three websites is gathered through web browsers.
To prevent copyright infringement, we initially consult the Robots Exclusion Protocol\footnote{\url{https://en.wikipedia.org/wiki/Robots.txt}} of the ExcelForum website.
The Robots Exclusion Protocol implemented by the ExcelForum website\footnote{\url{https://www.excelforum.com/robots.txt}} specifies that Bytespider is prohibited from accessing all directories, Mediapartners-Google is prohibited from crawling the xls files on the website, and all crawlers are prohibited from crawling the search.php directory.
Our crawler follows this Robots Exclusion Protocol of the website.
The crawler we use is not affiliated with Bytespider or Mediapartners-Google, and it does not crawl the search.php directory.
Though we follow the Robots Exclusion Protocol, considering potential licensing and copyright risks, we will refrain from engaging in any secondary distribution of the raw data.

\subsection{Maintenance Plan}
\label{app:maintenance}
The benchmark is licensed under CC BY-SA 4.0.
RUC KBReasoning group will maintain this benchmark in the long term.
We will continue to update the benchmark to fix labelling errors, instructions, or other necessary modifications, and all data will be versioned.
We also welcome all contributors interested in our benchmark.
If you have any questions or want to extend/augment/build on/contribute to the dataset, feel free to contact us via Github or E-mail (zeyaoma@ruc.edu.cn, zbhmint@bit.edu.cn).

% \clearpage
\section{Details of Data Collection}
\label{app:detail_data_collection}
In this section, we introduce our data collection process.
First, we present a concise overview of the four Excel websites that we use.
Subsequently, we introduce the detail of our data annotation process.
Finally, we show some data examples to intutively demonstrate the high quality of our data and the difference compared to previous benchmarks.

\subsection{Data Source}
\label{app:forum}
The data for our benchmark is obtained from four well-known Excel forums and blogs.
These websites contain diverse spreadsheet-related questions in the real world and high-quality answers.
Below is a brief introduction.

\textbf{ExcelForum}\footnote{\url{https://www.excelforum.com/}}
is one of the most popular spreadsheet forums in the world, containing 1 million threads, 5 million posts, and 1.3 million members as of May 2024.
The highest number of concurrent online users recorded was 7,174.
ExcelForum provides a detailed categorization of the spreadsheet-related problem.
The main categories are \textit{Excel General}, \textit{Excel Programming / VBA / Macros}, \textit{Excel Formulas \& Functions} and \textit{Excel Charting \& Pivots}.
The forum enforces strict rules to ensure the quality of threads, such as eliminating duplicate or ambiguous questions.
Moreover, the use of AI tools (e.g., ChatGPT, GPT-4) to create forum questions and answers is not permitted.
Although verifying AI-generated content can be challenging, the forum strives to minimize the presence of such content, which enhance the authenticity of the questions and the reliability of the answers.

\textbf{MrExcel}\footnote{\url{https://www.mrexcel.com/}}
is a widely recognized online platform that hosts more than 1 million discussion threads.
% Comparing to ExcelForum, MrExcel does not employ a detailed classification of questions, with more than 1 millon threads in a single category named \textit{Excel Questions}.
Unlike ExcelForum, MrExcel does not employ a detailed classification of questions. Instead, most questions are posted in one category called \textit{Excel Questions}, which contains over 1 million threads.
MrExcel adopts a number of rules to ensure the quality of threads and the authenticity of the answers (i.e., the forbidden of AI tools).

\textbf{ExcelGuru}\footnote{\url{https://excelguru.ca/}}
is an online platform designed to enhance one's understanding and proficiency in Excel and Power business intelligence (BI), which is an advanced function provided by Excel.
While the forum section of ExcelGuru is relatively small, the article section offers a number of excellent blogs and articles focused on Excel pivot tables and power queries.
Pivot tables and power queries are two powerful features that are used for data summarization, analysis, and reporting.
They extract meaningful insights from one table and create a new table to present them.
Each blog or article includes a complex question with a detailed solution, which is suitable for the data construction of our benchmark.

\textbf{Chandoo}\footnote{\url{https://chandoo.org/}}
is famous for its exceptional blogs, which are characterised by their superior quality and frequent updates.
On average, Chandoo publishes one blog per week, resulting in a grand total of over two thousand blogs.
While the publication of blogs is less frequent than forum posts, they offer well-organized questions, answers in spreadsheet format, and detailed instructional guides, making them more convenient for data formatting and a valuable addition to our benchmark.
The blogs on Chandoo covers a wide variety of topics, including \textit{VBA Macros}, \textit{Excel Challenges}, \textit{Formula Forensics}, \textit{Excel Howtos}, etc, which are all related to realistic user demands and advanced spreadsheet manipulation techniques.
Each blog contains a detailed solution and corresponding Excel files.

\subsection{Data Annotation}
\label{app:data_annotation}
We conduct three key annotation tasks during the construction of \bench, including (1) Data Selection, (2) Data Formatting, and (3) Test Case Generation.
We sequentially perform the three mentioned annotations and meticulously verify the data obtained from the previous annotation before each subsequent annotation to ensure the quality of the data.
% We first present the composition of our annotation team, followed by a concise description of the main procedures utilized for these three annotation tasks.

% \textbf{Annotation Team.}
% We employ a team of 20 individuals who specialize in Excel and have extensive experience in annotation.
% All team members hold a bachelor’s degree and are compensated in line with market rates.
% For data annotation, we allocate a predetermined quantity of data to annotators on a daily basis, and the allocation is modified in accordance with local regulations regarding work hours.
% For data validation, we hire two experienced validators with bachelor's degrees for the first round of quality checks.
% Moreover, two authors who hold master's degrees perform a secondary round of quality assessments to guarantee the high quality of data.

% \textbf{Annotation Process.}
% We sequentially perform the three mentioned annotations and meticulously verify the data obtained from the previous annotation before each subsequent annotation to ensure the quality of the data.

\begin{figure*}
  \centering
  \includegraphics[width=\textwidth]{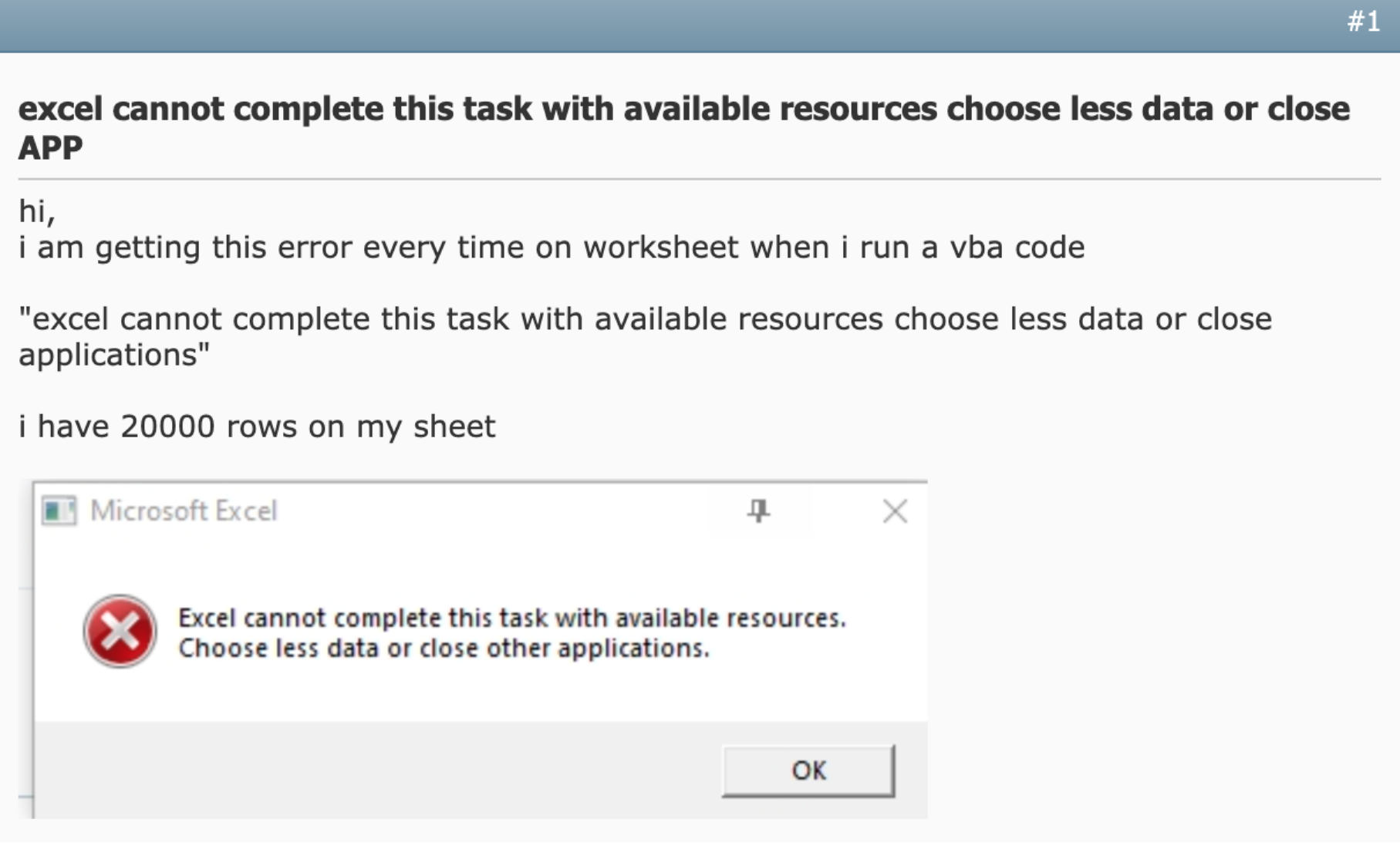}
  \caption{An example question about software usage, which are excluded from the dataset.} 
 \label{fig:excel_usage}
\end{figure*}

\begin{figure*}
  \centering
  \includegraphics[width=\textwidth]{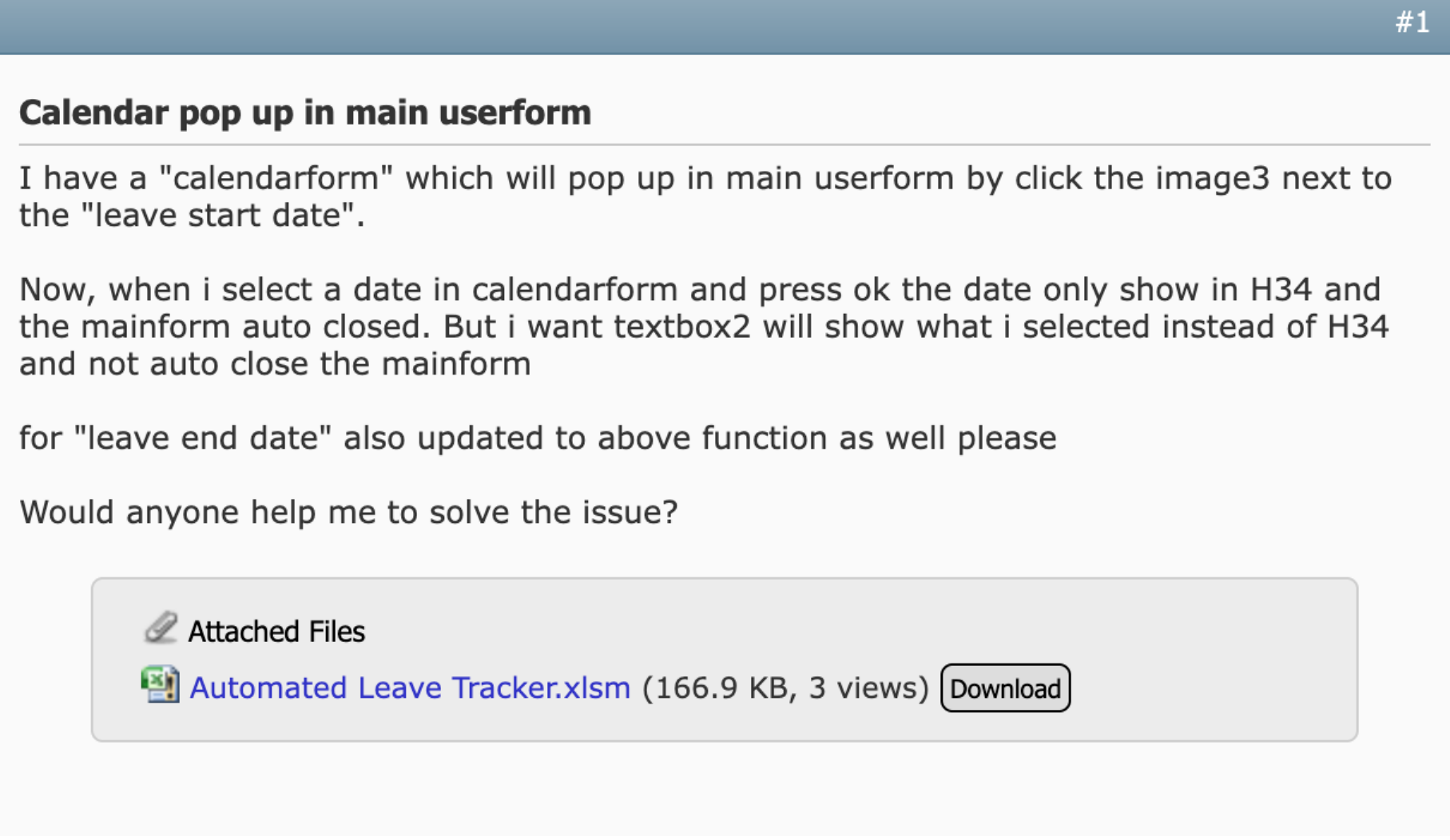}
  \caption{An example question related to component in Excel, which are excluded from the dataset.} 
 \label{fig:component_related}
\end{figure*}

\begin{figure*}
  \centering
  \includegraphics[width=\textwidth]{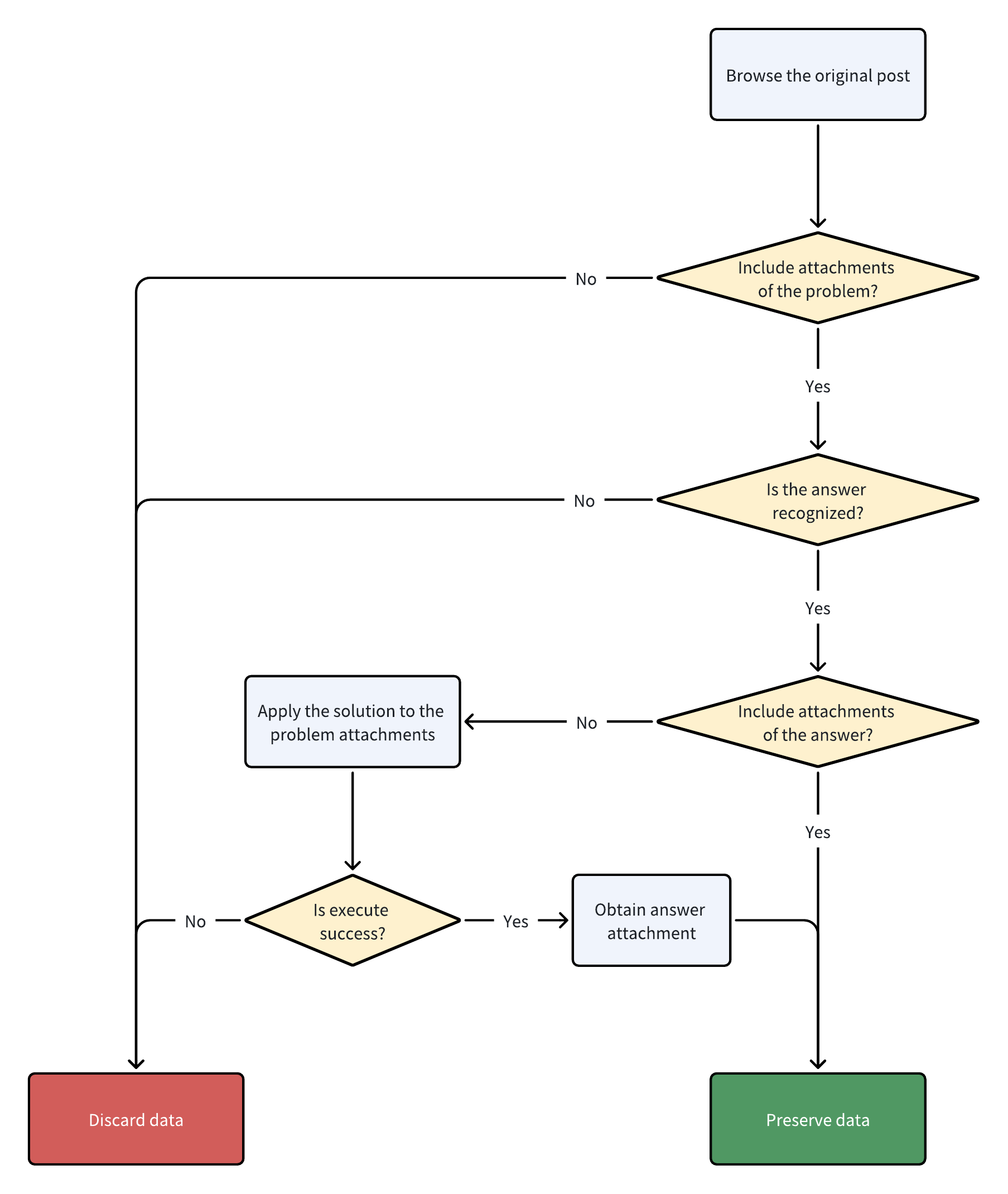}
  \caption{The flow chart of data selection process.} 
 \label{fig:data_selection}
\end{figure*}

\textbf{(1) Data Selection.}
We aim to select the data that is feasible and testable from a preliminary processed dataset.
The preliminary processed dataset is obtained by using keywords and LLM to exclude data that are unrelated to spreadsheet manipulation, such as software usage questions (See Figure~\ref{fig:excel_usage}) and questions related to components in Excel, such as button, userform, etc (See Figure~\ref{fig:component_related}).
The data selection pipeline is illustrated in Figure~\ref{fig:data_selection}.
First, we discard posts that do not include the spreadsheet files corresponding to the question, as these types of posts often focus on software usage or other questions that are not related to spreadsheet manipulation.
Subsequently, we eliminate posts that lack an acknowledged solution.
While certain posts may contain user-provided solutions, it is important to note that the questioner may not acknowledge or confirm whether these solutions adequately address their needs.
Finally, we try to obtain the answer spreadsheet files corresponding to the solution.
Some answerers provide both the solution and the answer spreadsheet files, and we can directly download these files as the initial answer.
Others may only offer the solution (e.g., formula, VBA code), and we need to manually apply these solutions in the attached spreadsheet files of the problem.
In this situation, we need to check whether the provided solution can be successfully executed.
Otherwise, we will discard the posts that contain solutions that cannot be executed.

\begin{figure*}
  \centering
  \includegraphics[width=\textwidth]{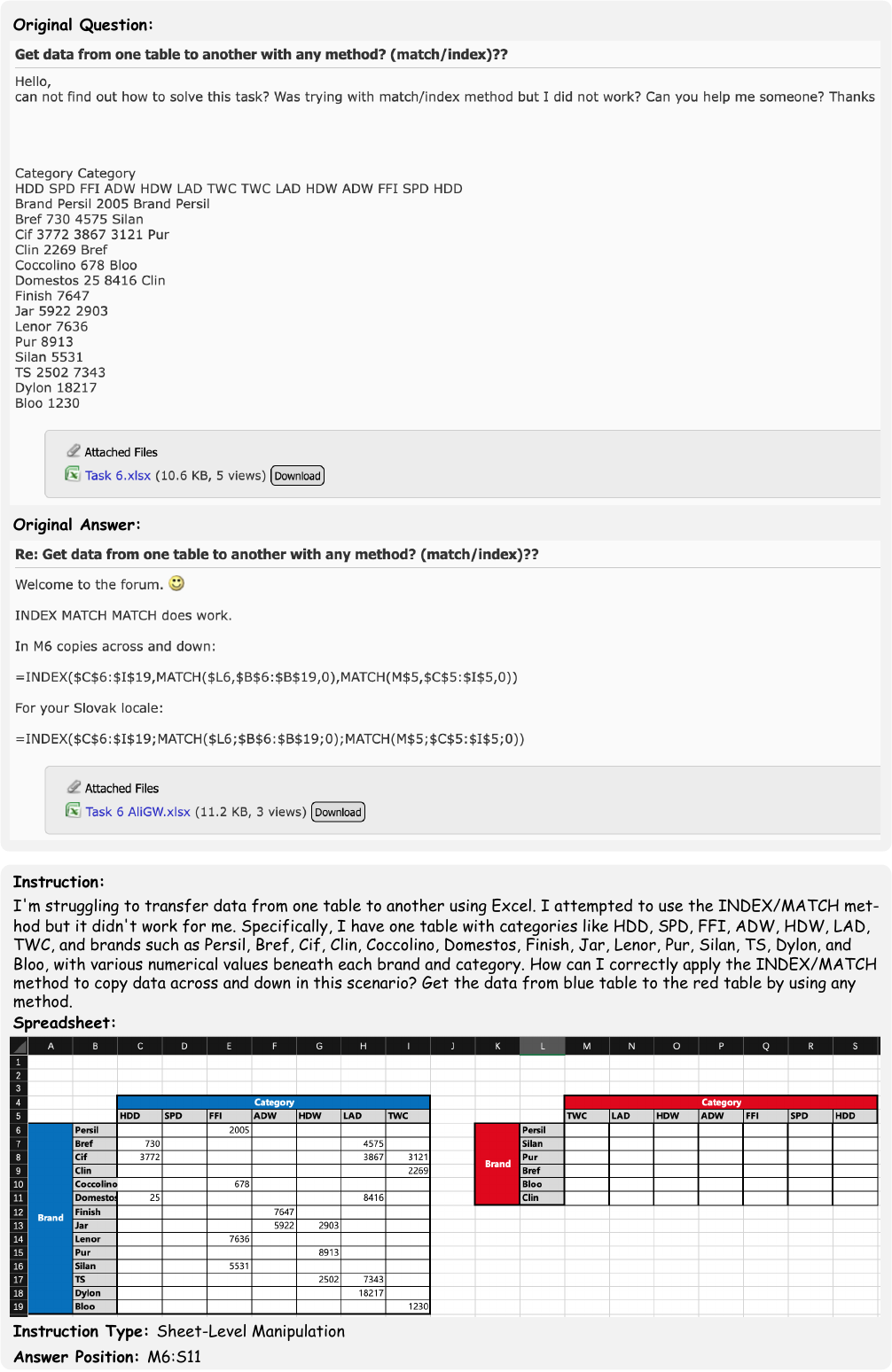}
  \caption{An example of the data formatting process. We revise the questions in the post and include supplementary details. We annotate the answer position based on the location applied by the formula. We apply the extracted solution from the post to the problem attachments (i.e., the attached spreadsheets)} 
 \label{fig:example_formatting}
\end{figure*}

\textbf{(2) Data Formatting.}
We aim to format the post into a self-contained question, an instruction type, an answer position, and a spreadsheet file corresponding to the question.
This data formatting process transforms the raw post into testable data and is feasible for our evaluation metrics (See Figure~\ref{fig:example_formatting} for an example).
The annotation team follows the instruction in Figure~\ref{tab:data_format} for the data formatting task.
First, we need to do the question verification to check whether the questions summarized by LLM are complete and accurate.
Second, we need to check the spreadsheet file corresponding to the question.
This involves removing some of the text-based responses and relocating the tabular data to the appropriate location.
Subsequently, we need to obtain the spreadsheet files corresponding to the solution, by applying the solution (e.g., formula, VBA code) to a specific position determined by the answerer.
Finally, we need to annotate the content of the solution, the answer position, and statistical information related to the spreadsheet content, including whether the spreadsheet file contains multiple tables and non-standard relational tables.

\begin{figure*}
  \centering
  \includegraphics[width=\textwidth]{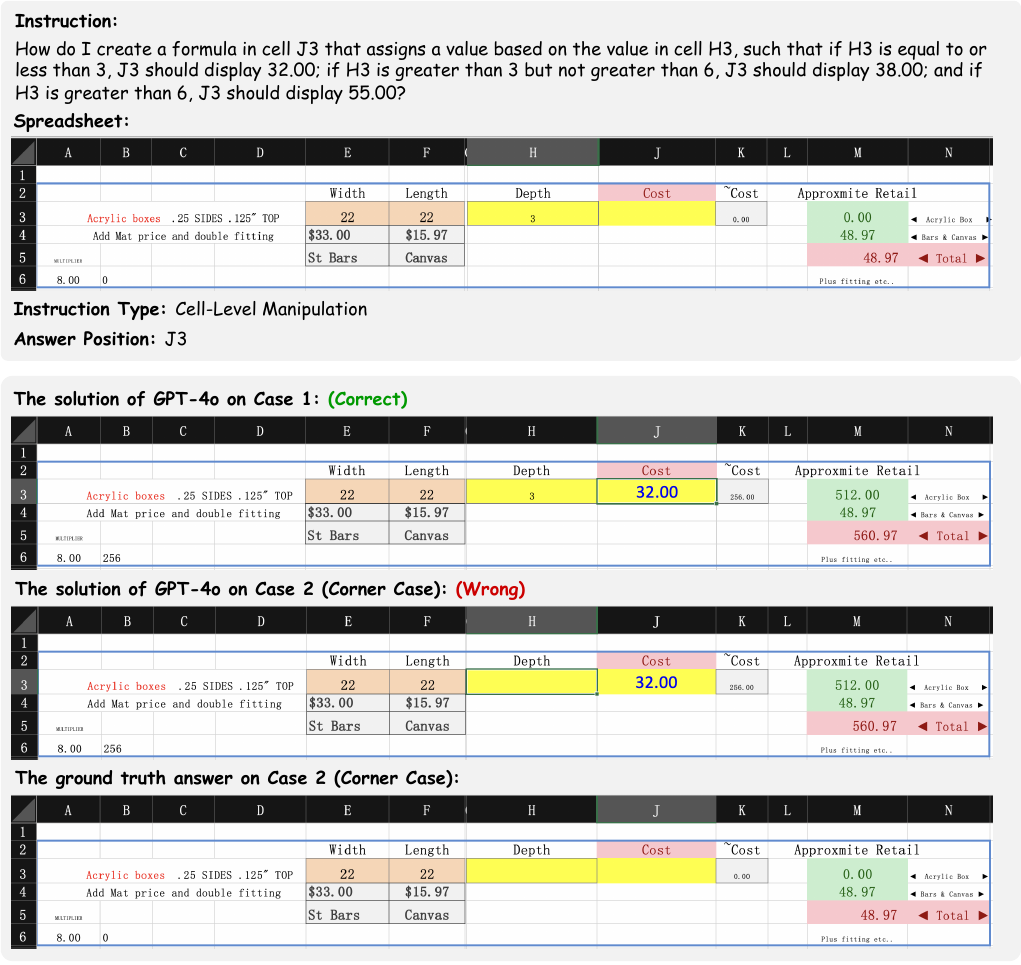}
  \caption{An example of the corner test case. In a practical situation, it is possible for cell H3 to be empty, and consequently, the value in cell J3 should also be empty. However, GPT-4o treats the value of empty cells as zero, leading to an incorrect result in cell J3.}
 \label{fig:example_corner}
\end{figure*}

\textbf{(3) Test Case Generation.}
We aim to generate two more test cases based on the original questions and the spreadsheet file.
The annotation manual is shown in Figure~\ref{tab:test_case_gen}.
Annotators are asked to modify the position applied by the solution to change the values of cells in answer position.
For solutions in formula format, the answer position change immediately once the position applied by the solution are modified.
For other situation (e.g., VBA code solution), annotators need to rerun the solution to obtain the updated answer.
Annotators must comprehend both the questions and the corresponding solution for each modification.
Additionally, they are instructed to carefully consider the exceptional scenarios in the questions and are encouraged to alter the cell value to simulate this exceptional situation (See Figure~\ref{fig:example_corner} for an example).

\subsection{Data Examples}
\label{app:data_examples}

As mentioned in Section~\ref{sec:statistic}, \bench incorporates complex questions based on real-world scenarios and diverse types of tables in spreadsheet files, compared to previous benchmarks.
In this section, we present four data examples in \bench to illustrate the attributes of real-world problems and the challenging of our benchmark.

\begin{figure*}
  \centering
  \includegraphics[width=\textwidth]{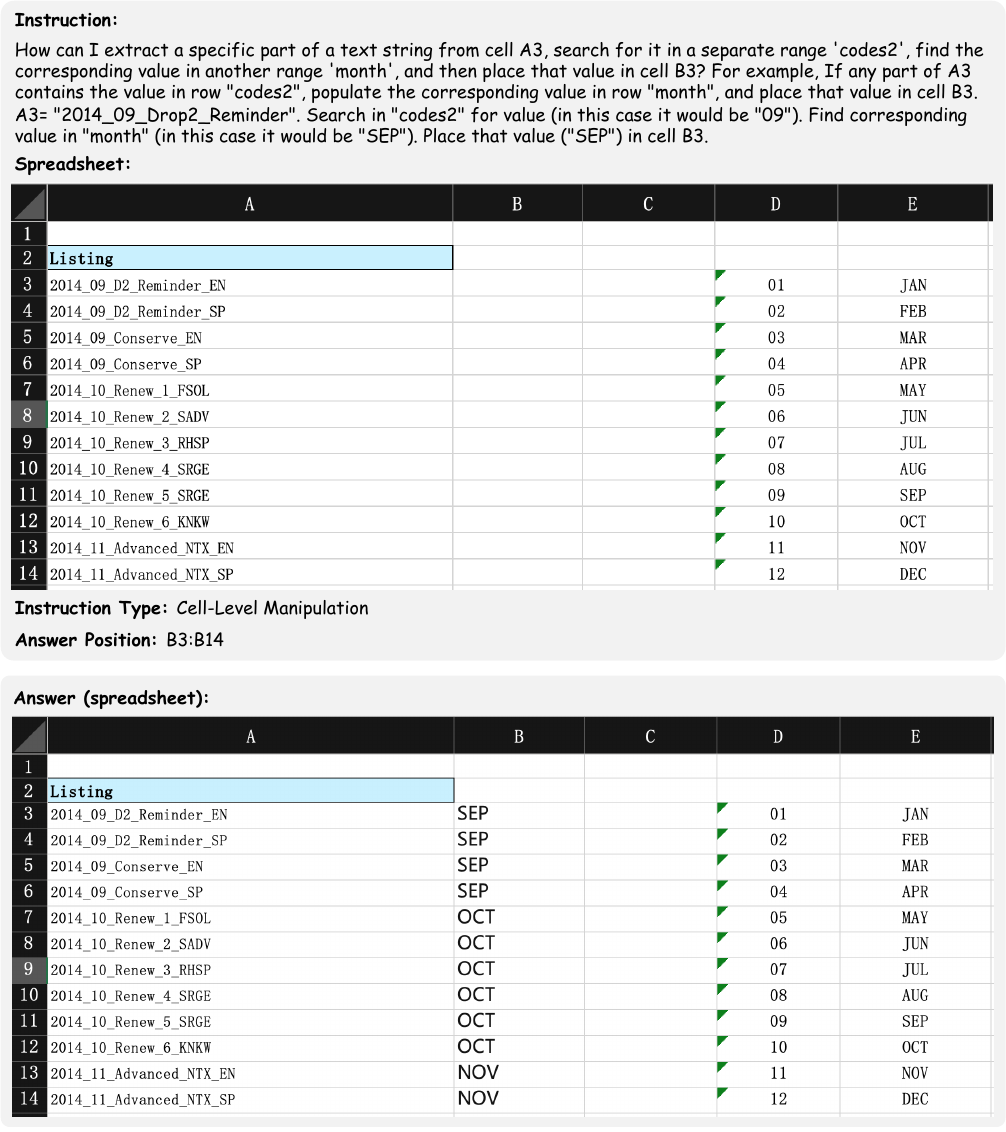}
  \caption{A cell-level manipulation example question involves manipulating a non-standard relational table (missing header on column D and column E).} 
 \label{fig:example_1}
\end{figure*}

Figure~\ref{fig:example_1} shows a cell-level manipulation data example which aims to extract a specific part of a text string from a column of cells.
The instruction contains the demand of the user and an example manipulating action of the question, which rarely occur in synthetic instructions of the previous benchmarks.
Furthermore, the table within the spreadsheet file is a non-standard relational table that lacks a complete table header.
The final result is required to be filled in the cells from B3 to B14, which ensure the uniqueness of the answer.

\begin{figure*}
  \centering
  \includegraphics[width=\textwidth]{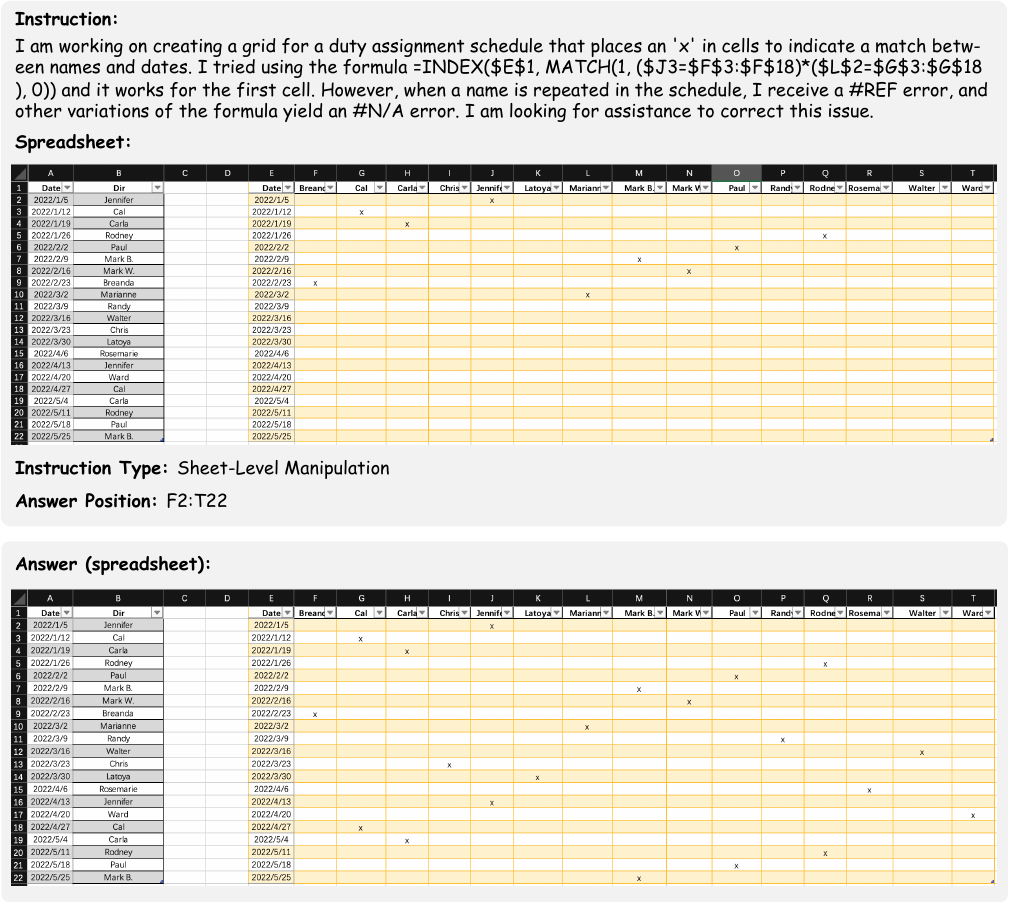}
  \caption{A sheet-level manipulation example question involves manipulating multiple tables in one sheet (first table from A1 to B22; second table from E1 to T22).} 
 \label{fig:example_2}
\end{figure*}

Figure~\ref{fig:example_2} shows a sheet-level manipulation data example aimed at creating a grid for a duty assignment schedule.
The instruction contains the demand of the user, the current formula solution and the error encountered by the user.
Besides, the spreadsheet file contains two relational tables in one sheet.
The final result should be filled in a two dimensional position (i.e., F2 to T22).

\begin{figure*}
  \centering
  \includegraphics[width=\textwidth]{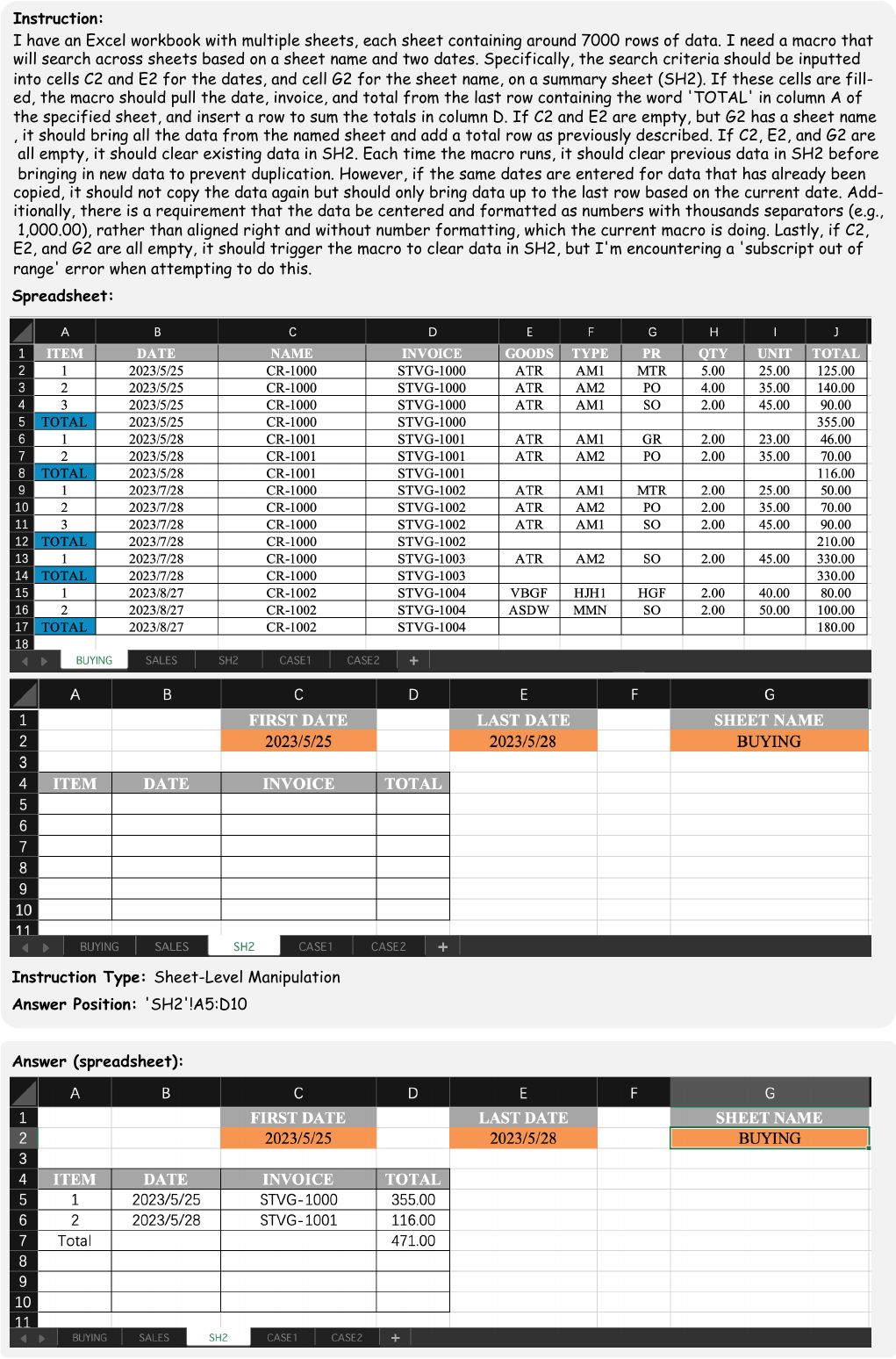}
  \caption{A sheet-level manipulation example question involves manipulating multiple tables on multiple sheets (first table in sheet named BUYING, from A1 to J17; second table in sheet named SH2, from A4 to D10).} 
 \label{fig:example_3}
\end{figure*}

Figure~\ref{fig:example_3} shows a sheet-level manipulation data example aims to search across sheets based on a sheet name and two dates (i.e., first date and last date).
The instruction contains a complex requirements, a detailed explanation, a manipulation example and the error encountered by the user.
Moreover, the spreadsheet file contains multiple sheets, including the table to be manipulated (e.g., sheets named BUYING or SALES) and examples of possible answers (e.g., sheets named CASE1 or CASE2).
The final result should be inserted into a two-dimensional location on the sheet named SH2.

\begin{figure*}
  \centering
  \includegraphics[width=\textwidth]{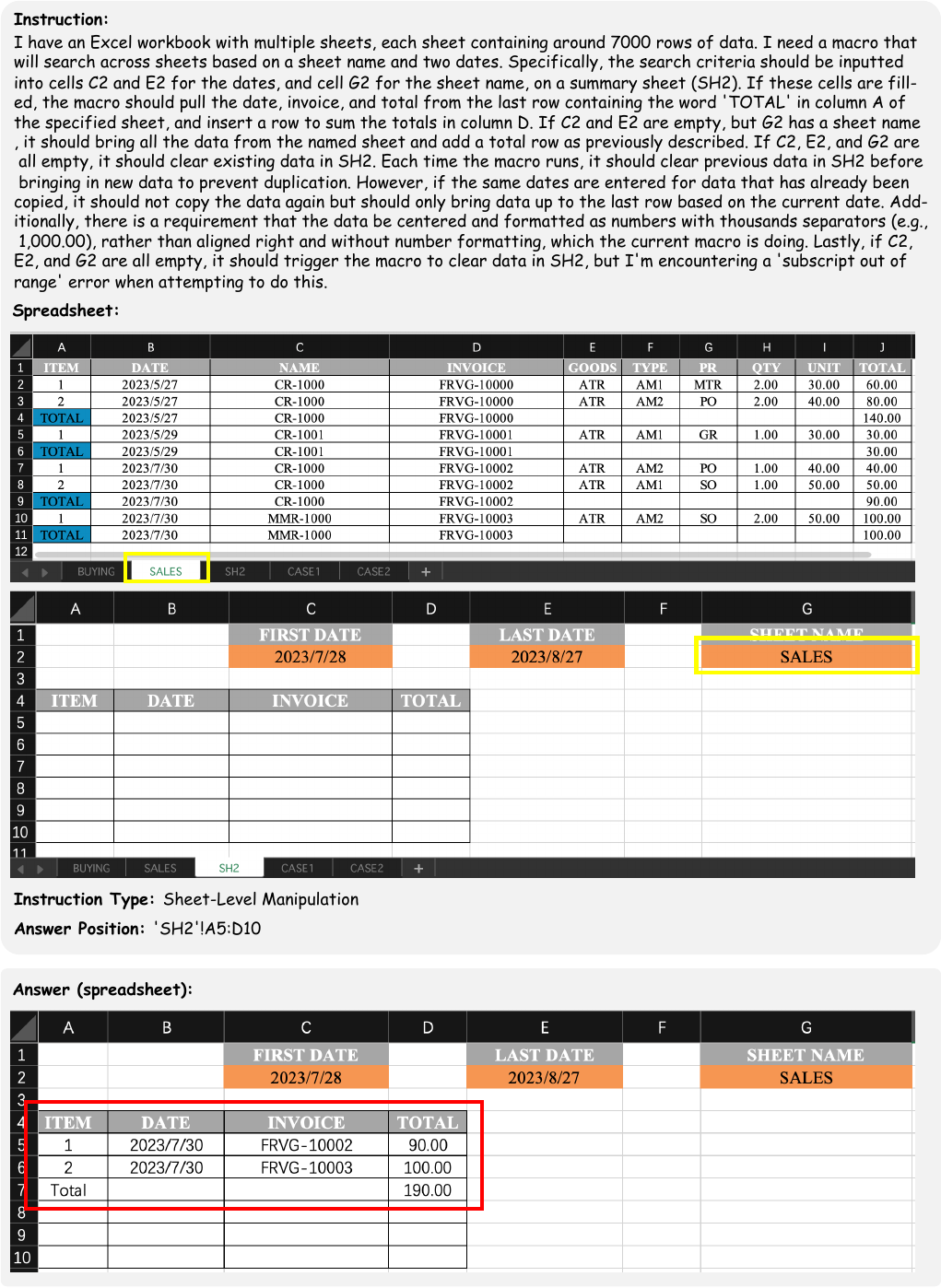}
  \caption{A test case modified from the question in Figure~\ref{fig:example_3}.} 
 \label{fig:example_4}
\end{figure*}

Figure~\ref{fig:example_4} shows a test case modified from the question in Figure~\ref{fig:example_3}.
This test case is constructed following the user's requirements in the instruction.
The user expects to obtain a result that will search within a specific sheet name (cell G2).
Only the rows with a date falling between the first date (cell C2) and last date (Cell E2) should be returned.
Therefore, we modify the value of the sheet name (cell G2) to construct a new test case, as highlighted in yellow.
This test case yields search results in the SALES sheet, rather than the original case that searches in the BUYING sheet.
The values in the answer position also change, as highlighted in red.
By modifying the sheet name, we can alter the sheet that is being searched, enabling a more comprehensive assessment of the solution's effectiveness.

% \clearpage
\section{Details of Result Inference}
\label{app:detail_exp}
In this section, we introduce the inference details and human evaluation settings of our main experiments.
Subsequently, we carry out supplementary experiments on our multi-round inference setting and present a detailed analysis of our evaluation metrics.

\subsection{Deployment Environment and Model Information}
The baseline models are divided into five types, including \textit{TableQA models}, \textit{Open-source code models}, \textit{Open-source general models} and \textit{Spreadsheet-specific methods or products}.
For open-source models that need to be deployed on our own, we utilize \textit{PyTorch}, \textit{transformers}, and \textit{vLLM} to load the model.
We conduct experiments and obtain inference results on an Ubuntu 22.04 server with graphic cards that contain 8 NVIDIA A100 SXM 80GB GPUs.
For spreadsheet-specific methods or products, we evaluate within Google Sheets\footnote{\url{https://www.google.com/sheets/about/}} and Microsoft Excel\footnote{\url{https://www.microsoft.com/microsoft-365/excel}} on Windows 10 and macOS.
For the detailed information of each baseline model, please refer to the website url in Table~\ref{tab:model_info}.

\input{table/model_info}

\subsection{Details of Experimental Setup}
All the baseline LLMs share a unified hyper-parameter (temperature to 1 and top\_p to 1) on two inference settings.
The prompt on single round setting and multiple round setting are shown in Figure~\ref{tab:single_prompt} and Figure~\ref{tab:multiple_prompt}, respectively.
For spreadsheet-specific methods or products including SheetCopilot and Copilot in Excel, two of the authors manually upload the spreadsheet files to Google Sheets or the cloud storage of Excel to evaluate the corresponding instructions.
As a result of the financial and time costs, we limit our assessment to fifty instructions and one test case per instruction.

% \clearpage
\section{Additional Experiments}

\subsection{Analysis of Inference Settings}
The main result presented in Table~\ref{tab:main_result} reveals an intriguing observation: the performance of GPT-4o in multi-round setting is slightly lower to that in single-round setting, which is contrary to the result of other models.
In this section, we perform comprehensive experiments to determine the influence of various components in the multi-round setting on the outcomes.
Furthermore, we perform multiple times of experiments for each inference setting to ensure the statistical significance of the results.
Below is the introduction of four inference settings:

\input{table/ablation_setting}
\label{app:infer_setting}

(1) \textit{Single + 5 Rows}:
This setting is identical to the single-round setting that we employed in the main experiments.
We add five rows of the spreadsheet file in pandas dataframe format to the prompt and obtain the direct code solution of LLM in the single round setting.

(2) \textit{Multiple + Execution Feedback + 5 Rows}:
Compared to setting (1), LLMs is able to refine the code solution within five rounds if the solution fails to execute.
The executor will provide the error traceback to the LLM to start the conversation of next round.

(3) \textit{Multiple + ReAct + Execution Feedback}:
Compared to setting (2), we introduce the ReAct mechanism to prompt LLMs to autonomously retrieve the content of the spreadsheet file, while excluding the five rows from the prompt.

(4) \textit{Multiple + ReAct + Execution Feedback + 5 Rows}:
This setting is identical to the multi-round setting that we employed in the main experiments.
Based on the setting (3), we include the five rows of the spreadsheet file in the prompt.
% Compared to setting (3), we also include the ReAct mechanism to prompt LLMs, while including the five rows of the spreadsheet file in the prompt.

The experiments are performed on a sample of 200 data points (21.9\% of the entire benchmark) and we use GPT-3.5 and GPT-4o for evaluation.
We conduct four runs for each setting and calculate the average performance, as illustrate in Table~\ref{tab:ablation_setting}.
The following findings have been discovered:

\textbf{1. LLMs benefit from multi-round setting.}
As shown in Table~\ref{tab:ablation_setting}, the performance of GPT-3.5 and GPT-4 improves in three multi-round settings when compared to the single-round setting.
This may be attributed to either the ReAct mechanism, the feedback of code execution, or both, which we will determine later.
Besides, the performance of GPT-4o on multi-round setting is higher than on single-round setting, which is inconsistent with the main result in ~\ref{tab:main_result}.
This is due to our practice of conducting four times of experiments and calculating the average result, which enhances the stability and reliability of our findings.
We also enforce strict alignment between single-round and multi-round prompts by deleting one sentence that ask the LLM to answer carefully in the single-round prompt.

\textbf{2. LLMs perform worse when adding ReAct mechanism.}
Comparing settings (2), (3), and (4), we find that using only the feedback from code execution achieves the highest performance.
While the ReAct mechanism provide the model with an additional way to obtain spreadsheet information (instead of only 5 rows from the prompt), they are unable to effectively utilize this additional data.
While the ReAct mechanism provides the model with an additional way to obtain spreadsheet information (beyond just the 5 rows from the prompt), it is unable to effectively utilize this additional data. Furthermore, existing LLMs lack the ability to make flexible decisions about whether to proceed with reading the content of the spreadsheet or which additional sections should be read. For instance, the model may continue to read the content of the first five rows of the spreadsheet even though this information has already been provided in the prompt. This phenomenon demonstrates that the current state-of-the-art LLM still lacks spreadsheet understanding and reasoning ability.

\subsection{Analysis of Impact Factors}

\begin{figure*}
  \centering
  \includegraphics[width=0.9\textwidth]{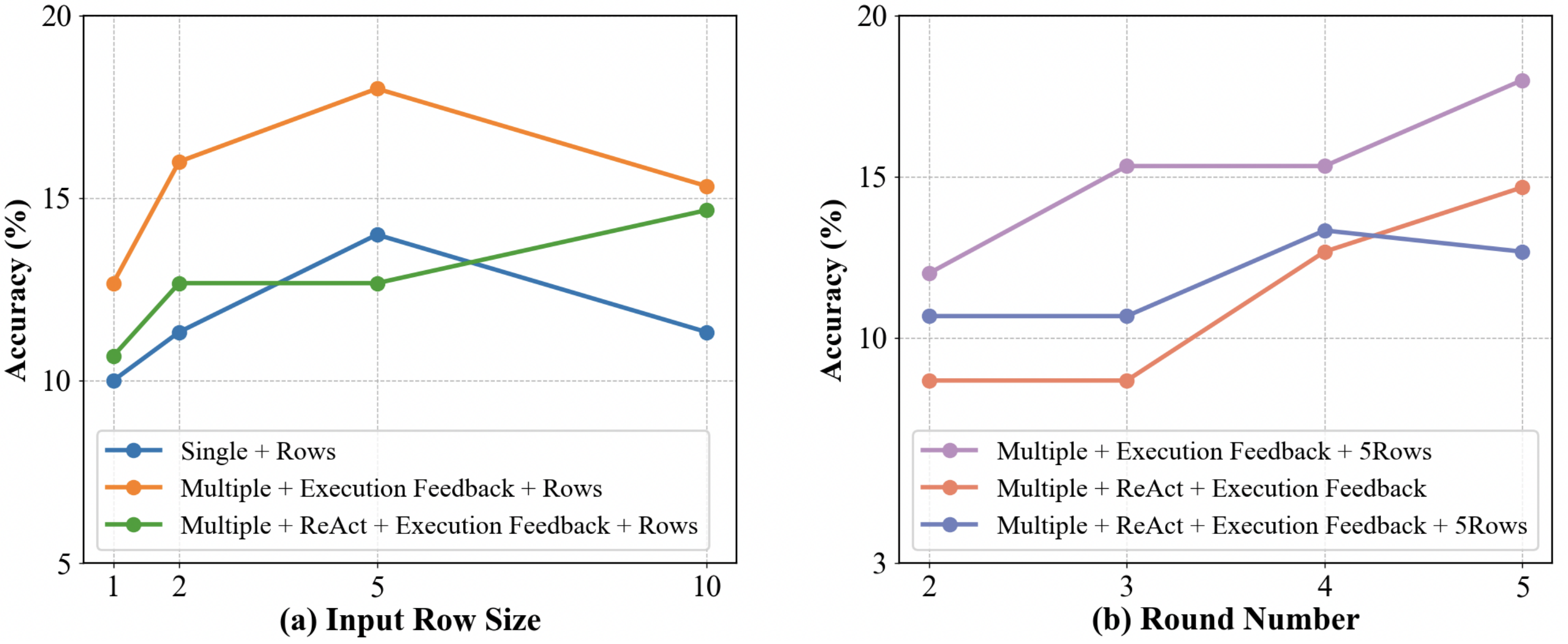}
  \caption{The impact of input row size and round number on GPT-4o across different inference settings. Accuracy represents the value of the overall soft restriction.} 
 \label{fig:ablation_rows_rounds}
\end{figure*}

Based on the four inference settings introduced in Appendix~\ref{app:infer_setting}, we conduct an analysis of two key factors in these settings, including the the number of rows provided in the prompt and the round number in multiple round setting.

The experiments are performed on a sample of 50 data points.
For setting (1), (2) and (4), we modify the number of rows specified in the prompt to 1, 2, 5, and 10 rows, and analyze the impact on the performance of GPT-4o.
For setting (2), (3), and (4), we modify the number of rounds of the interactive inference environment.
The following findings have been discovered:

\textbf{1. The performance of LLMs benefits from a limited number of rows.}
As illustrated in Figure~\ref{fig:ablation_rows_rounds}(a), the performance of GPT-4o improves across all three settings as the row size increases from 1 row to 5 rows.
However, in two of the settings, the performance declines as the row size increases to 10 rows.
This phenomenon is consistent with our observation in the result of Appendix~\ref{app:infer_setting}.
LLMs fail to process and comprehend a relatively large number of rows, thus requiring improvements in LLMs for understanding tabular data.

\textbf{2. The performance of LLMs increases with the number of rounds.}
As illustrated in Figure~\ref{fig:ablation_rows_rounds}(b), GPT-4o shows an overall upward trend in performance as the number of rounds increases in all three settings.
This result is inconsistent with the result in Figure~\ref{fig:impact_factor}(b), as we conduct the experiments multiple times and obtain an average result, which is much more stable.
We also find that setting (3) is comparable to setting (4) when the number of rounds is four and five.
This phenomenon illustrates that the current LLMs are incapable of effectively retrieving information beyond the first five rows of the spreadsheet in the ReAct mechanism.

\subsection{Analysis of Evaluation Metrics}
As our evaluation metrics are based on exact match of the cells in the answer position, we conduct a human evaluation to verify the reliability of the metrics.
Specifically, we sample 50 instructions from the entire benchmark with three test cases of each and sample the corresponding inference result of GPT-4o.
We follow DS-1000~\cite{lai2023ds1000} and report the following four indexes:

(1) \textit{Test Case Level False Discovery Rate}:
Among all the test cases that successfully pass our automated evaluation, none of them are deemed incorrect by our annotator.

(2) \textit{Test Case Level False Omission Rate}:
Among all the test cases that fail our automatic evaluation, 3.8\% of them are deemed correct by our annotator.

(3) \textit{Instruction Level False Positive Percentage}:
Among all the instructions, none of them have any incorrect sample predictions that pass our automatic metric.

(4) \textit{Instruction Level False Negative Percentage}:
Among all instructions, 4\% instructions contain at least one correct sample prediction that fails to pass our automatic metric.

\begin{figure*}
  \centering
  \includegraphics[width=\textwidth]{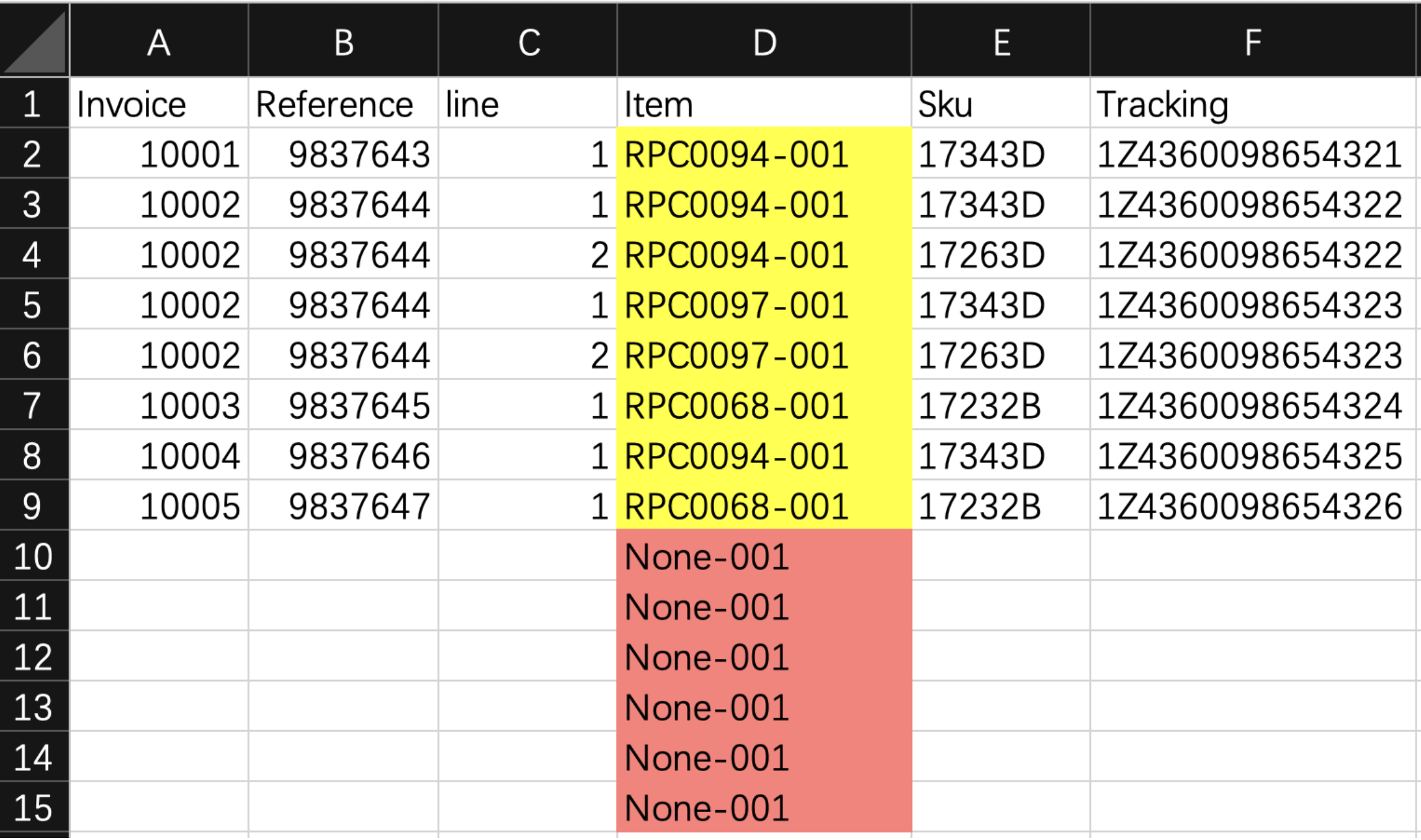}
  \caption{A misclassified test case. The code solution successfully generates the correct answer (highlighted in yellow). However, additional content (highlighted in red) is also generated, resulting in incorrect judgments by our evaluation metrics.} 
 \label{fig:misclassify_1}
\end{figure*}

\begin{figure*}
  \centering
  \includegraphics[width=\textwidth]{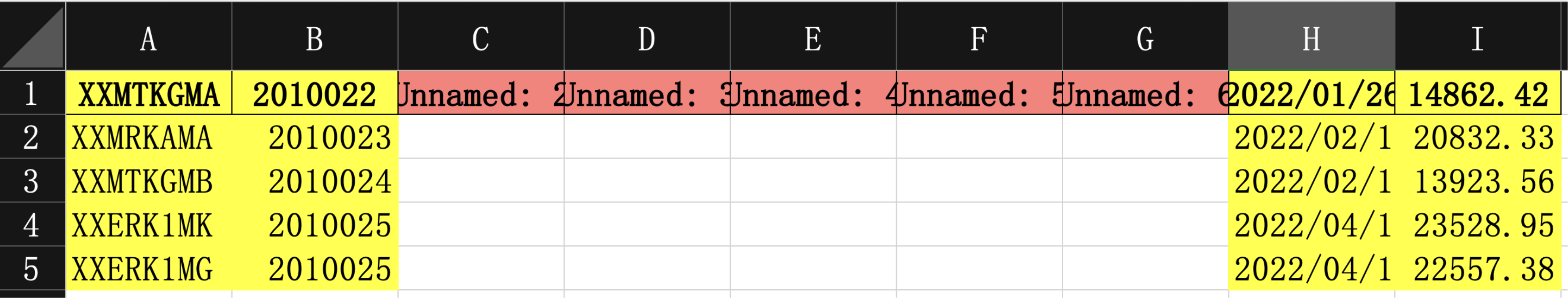}
  \caption{A misclassified test case. The code solution successfully preserve the target rows (highlighted in yellow). However, additional table headers (highlighted in red) are also generated, resulting in incorrect judgments by our evaluation metrics.} 
 \label{fig:misclassify_2}
\end{figure*}

Our evaluation metrics do not misclassify accurate results as inaccurate ones, as we achieve a 0\% error rate in indices (1) and (3).
It may misclassify a small proportion of correct test cases to incorrect ones.
This occurs because the code solution generated by LLMs may produce additional content beyond the intended answer (See Figure~\ref{fig:misclassify_1} and Figure~\ref{fig:misclassify_2}).
In general, our evaluation metric is reliable, as it accurately identifies correct results and only misclassifies a small fraction of correct test cases as incorrect.

\clearpage

% Case study: Example 1
\begin{figure*}
  \centering
  \includegraphics[width=\textwidth]{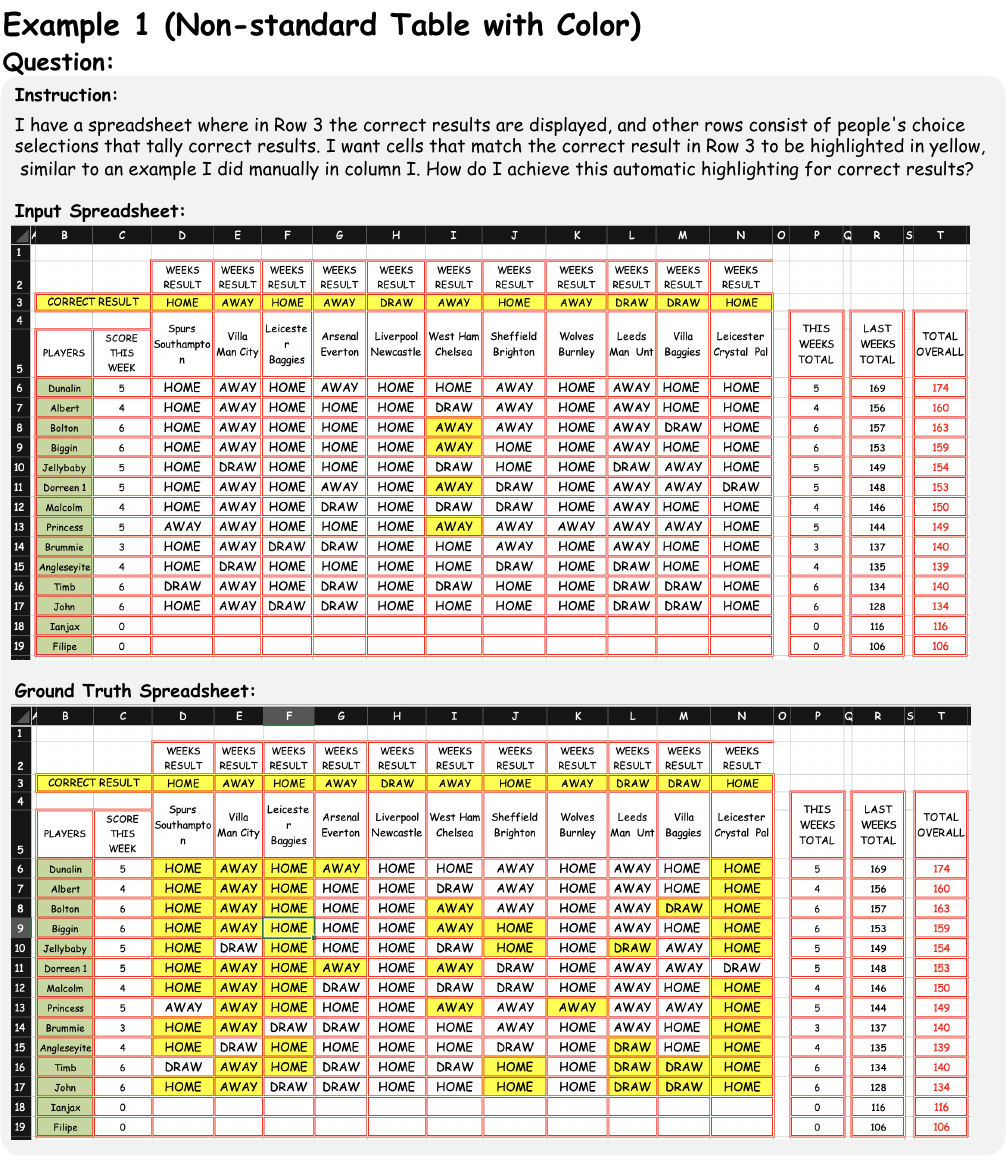}
  \caption{The question of Example 1 with non-standard table and color.} 
 \label{fig:case_study_1-1}
\end{figure*}

\begin{figure*}
  \centering
  \includegraphics[width=\textwidth]{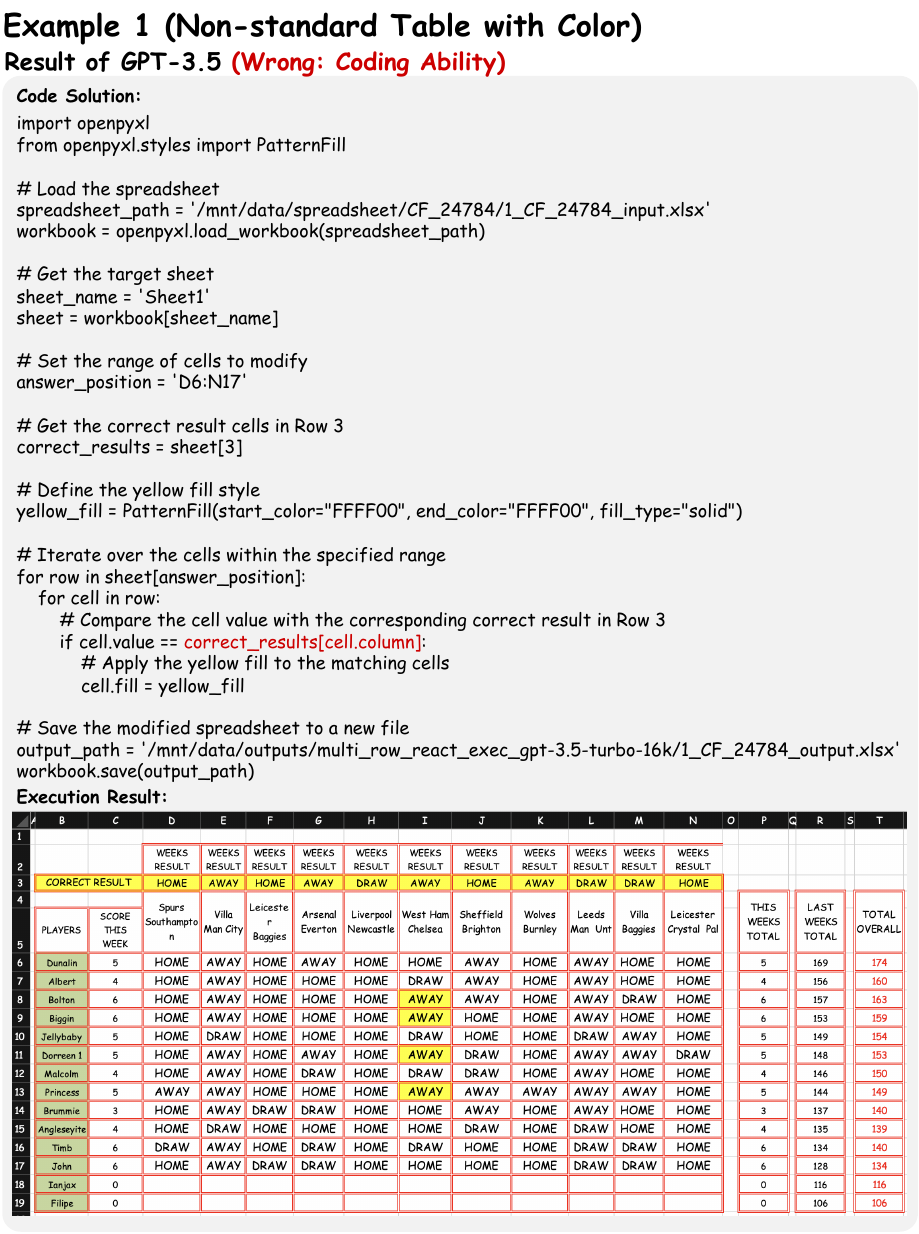}
  \caption{GPT-3.5 fails on Example 1 because of its lack of coding ability.}
 \label{fig:case_study_1-2}
\end{figure*}

\begin{figure*}
  \centering
  \includegraphics[width=\textwidth]{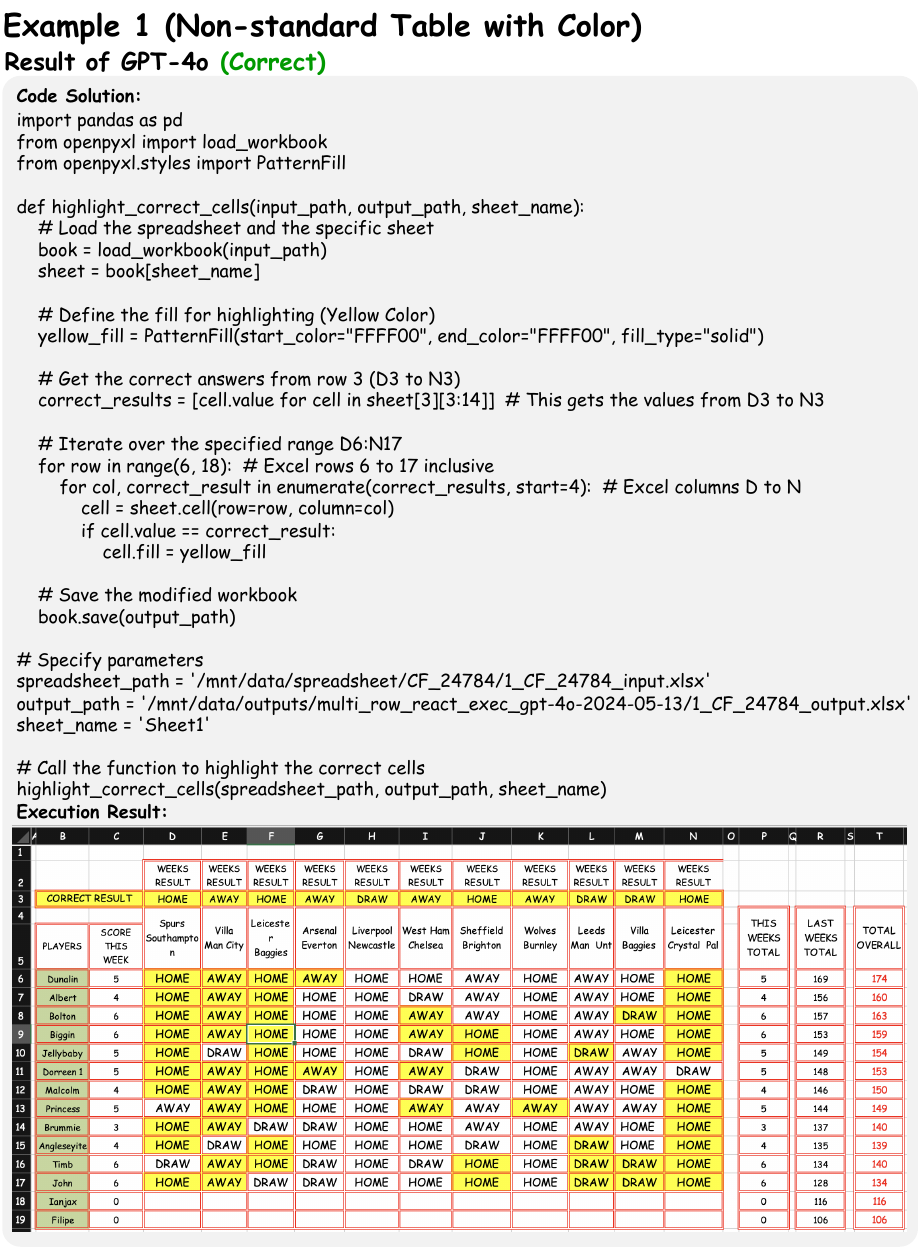}
  \caption{GPT-4o gets the right answer on Example 1.} 
 \label{fig:case_study_1-3}
\end{figure*}

% Case study: Example 2
\begin{figure*}
  \centering
  \includegraphics[width=\textwidth]{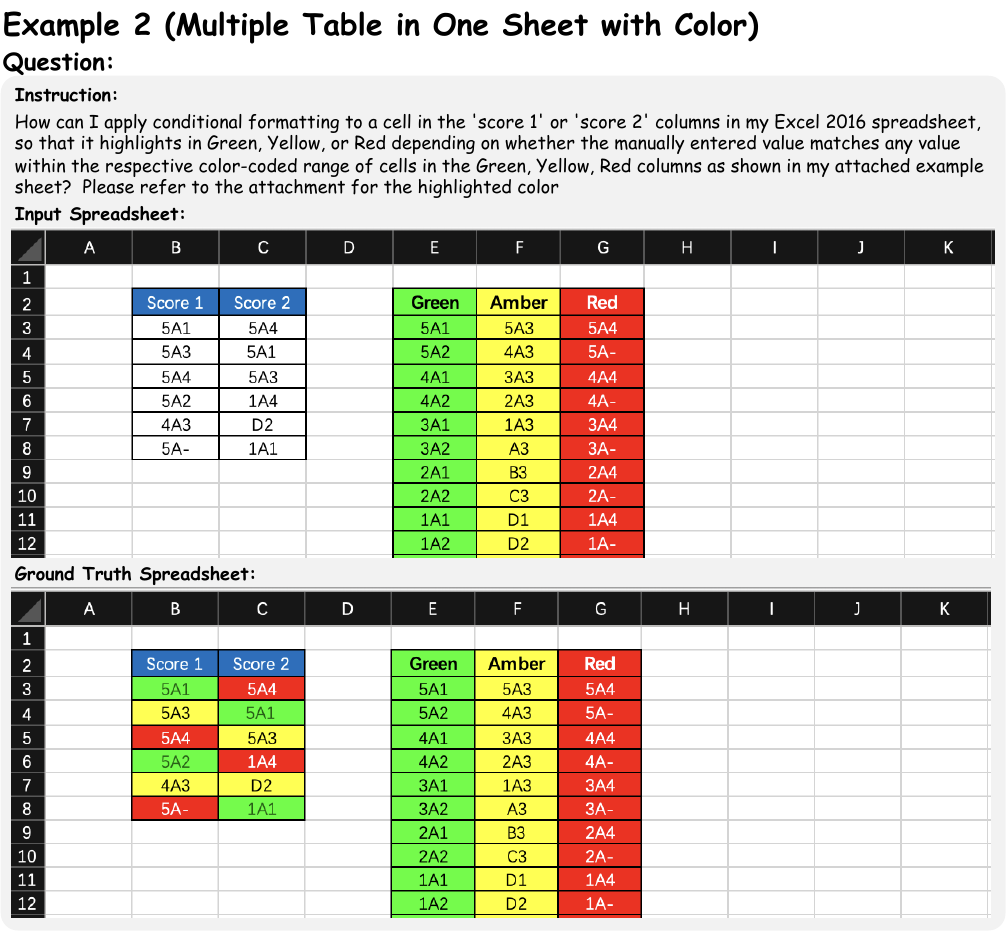}
  \caption{The question of Example 2 with multiple table and color.} 
 \label{fig:case_study_2-1}
\end{figure*}

\begin{figure*}
  \centering
  \includegraphics[width=\textwidth]{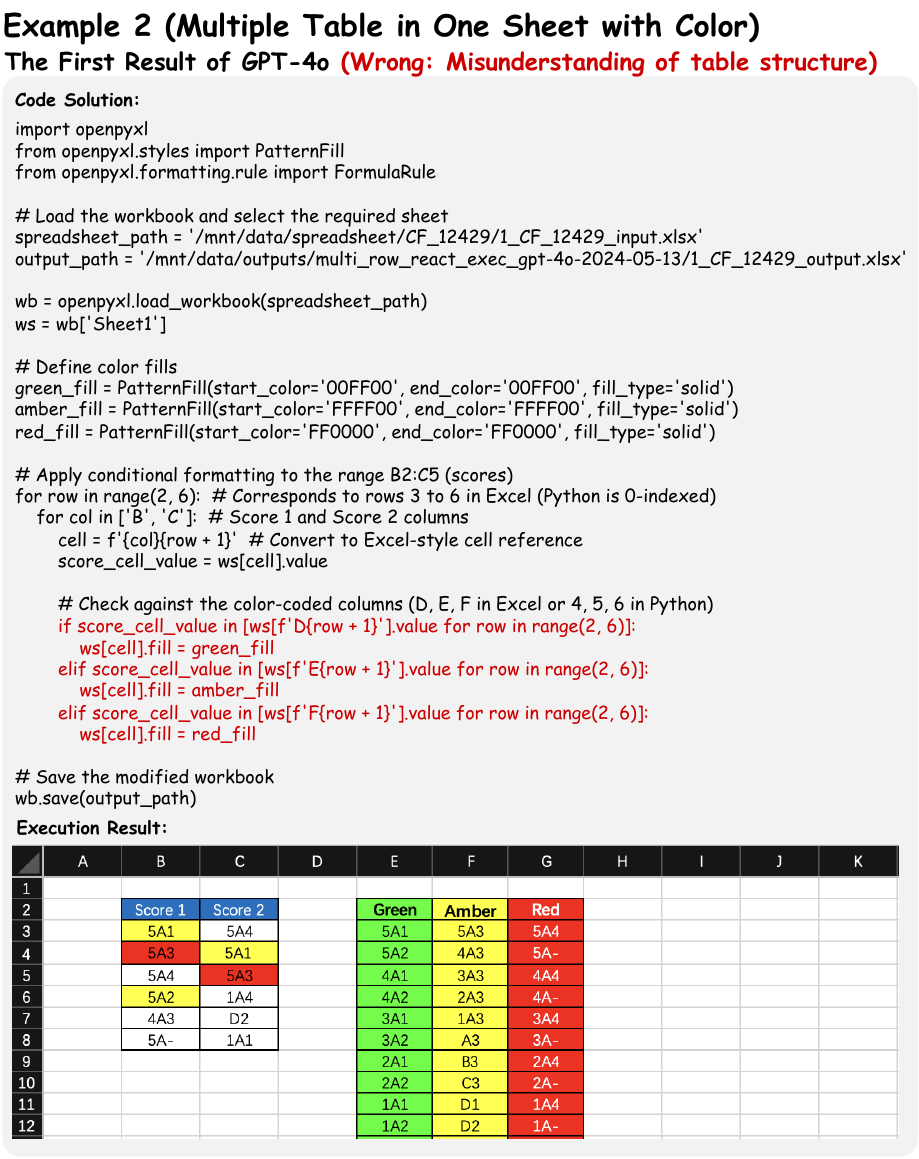}
  \caption{GPT-4o fails on Example 2 because of misunderstanding of table structure.}
 \label{fig:case_study_2-2}
\end{figure*}

\begin{figure*}
  \centering
  \includegraphics[width=\textwidth]{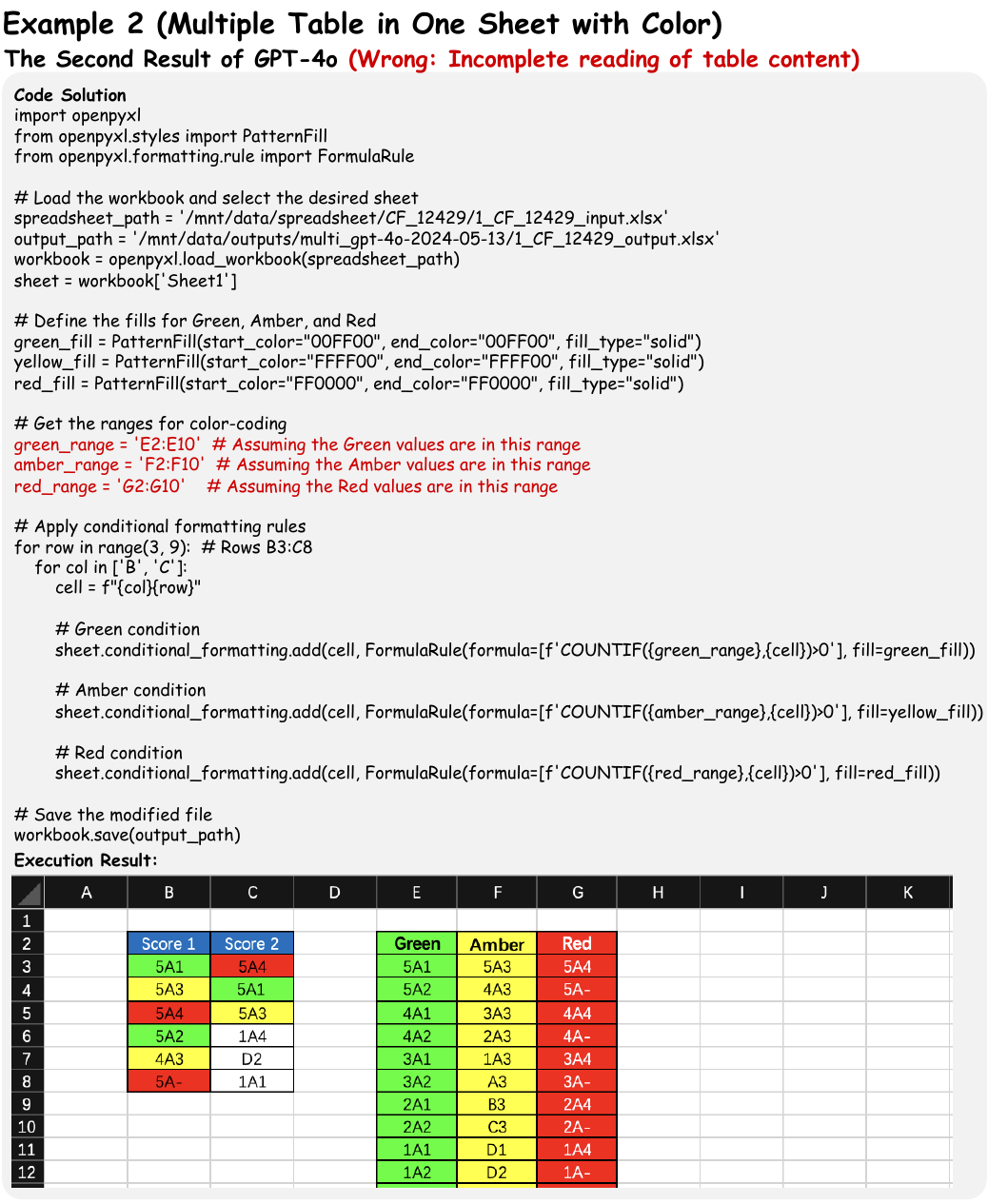}
  \caption{GPT-4o fails on Example 2 because of incomplete reading of table content.} 
 \label{fig:case_study_2-3}
\end{figure*}

% data example

\input{table/data_format}

\input{table/test_case_gen}

\input{table/prompt_single}

\input{table/prompt_multiple}

%% file: table/model_info.tex
\begin{table}[h]
  \caption{Information of baseline models.}
  \label{tab:model_info}
  \centering
  \resizebox{\linewidth}{!}{
  \begin{tabular}{l|r|l}
    \toprule
    Model & Size & Website Url \\
    \midrule
    TaPEx & 400M & \url{https://huggingface.co/microsoft/tapex-large-finetuned-wtq} \\
    TaPas & 340M & \url{https://huggingface.co/google/tapas-large-finetuned-wtq} \\
    Binder & * & \url{https://github.com/xlang-ai/Binder} \\
    \midrule
    CodeQwen1.5 & 7B & \url{https://huggingface.co/Qwen/CodeQwen1.5-7B-Chat} \\
    DeepseekCoder & 33B & \url{https://huggingface.co/deepseek-ai/deepseek-coder-33b-instruct} \\
    \midrule
    Mixtral-8x7B & 47B & \url{https://huggingface.co/mistralai/Mixtral-8x7B-Instruct-v0.1} \\
    Llama-3 & 70B & \url{https://huggingface.co/meta-llama/Meta-Llama-3-70B-Instruct} \\
    \midrule
    GPT-3.5 & * & \url{https://platform.openai.com/overview} \\
    GPT-4 & * & \url{https://platform.openai.com/overview} \\
    \midrule
    SheetCopilot & * & \url{https://github.com/BraveGroup/SheetCopilot} \\
    Copilot in Excel & * & \url{https://www.microsoft.com/microsoft-copilot/} \\
    \bottomrule
  \end{tabular}}
\end{table}

%% file: table/ablation_setting.tex
\begin{table}[t]
  \caption{Impact of different inference settings on the performance of GPT-3.5 and GPT-4o (\%).}
  \label{tab:ablation_setting}
  \centering
  \resizebox{\linewidth}{!}{
  \begin{tabular}{l|l|c|c}
    \toprule
    Model & Inference Setting & Soft Restriction & Hard Restriction \\
    \midrule
    \multirow{4}{*}{GPT-3.5} & Single + 5 Rows & 5.38 ($\pm$2.19) & 4.00 ($\pm$1.55) \\
    & Multiple + Execution Feedback + 5 Rows & \textbf{6.58} ($\pm$1.42) & \textbf{4.50} ($\pm$1.50) \\
    & Multiple + ReAct + Execution Feedback & 5.50 ($\pm$0.33) & \underline{4.38} ($\pm$2.12) \\
    & Multiple + ReAct + Execution Feedback + 5 Rows & \underline{6.04} ($\pm$0.59) & 4.25 ($\pm$0.75) \\
    \midrule
    \multirow{4}{*}{GPT-4o} & Single + 5 Rows & 12.79 ($\pm$1.71) & 10.63 ($\pm$1.88) \\
    & Multiple + Execution Feedback + 5 Rows & \textbf{16.08} ($\pm$1.08) & \textbf{13.50} ($\pm$1.00) \\
    & Multiple + ReAct + Execution Feedback & 13.54 ($\pm$1.27) & 10.88 ($\pm$1.12) \\
    & Multiple + ReAct + Execution Feedback + 5 Rows & \underline{14.67} ($\pm$1.77) & \underline{11.75} ($\pm$1.25) \\
    \bottomrule
  \end{tabular}}
\end{table}

%% file: table/data_format.tex
\begin{figure*}
    \centering
    \begin{mdframed}[linecolor=black, linewidth=1pt, backgroundcolor=gray!10]
    \small

\textit{Annotation manual of data formatting task}:\\

[1. Question Verification]\\
1. You need to compare the original post text with the question regenerated by the large language model based on the post content. If the question generated by the large model misses some information from the original post, you need to add this missing information to the question and fill it in under the column named ``Revised Question''.\\
2. You need to check if the attachment corresponding to the question contains any additional textual information that can be added to the question, and add it to the question when appropriate.\\

[2. Spreadsheet Verification]\\
1. You need to check if the spreadsheet provided by the asker already contains text-based answers in the target locations. If the text-based answers are identical to the results after the application of the solution, you need to delete some of the text-based answers and only keep a portion as examples.\\
2. You need to check if the data in the spreadsheet is too far from the A1 position, for instance, if the data in the spreadsheet starts from G200. In this case, you need to move the data in the spreadsheet closer to the A1 position for easier evaluation later on.\\

[3. Solution Verification and Application]\\
1. If the post provides an Excel file containing the answers, you need to check if the content in the file answers the asker's question.\\
2. If the post does not provide an Excel file with the answers, you need to apply the solution provided by the author (e.g., formulas or VBA code) to a specific location in the spreadsheet, which is generally specified by the responder.\\

[4. Data Annotation]\\
1. Solution Annotation: Record the solutions, such as formulas or VBA code.\\
2. Answer Position Annotation: Record the cells where the solutions are applied, such as A1:B10.\\
3. Spreadsheet Content Annotation: Based on the examples of multi-tables and non-standard tables we provide, determine if there are multi-tables and non-standard tables in the file.

    \end{mdframed}
    \caption{The annotation manual of data formatting task.}
    \label{tab:data_format}
\end{figure*}

%% file: table/test_case_gen.tex
\begin{figure*}
    \centering
    \begin{mdframed}[linecolor=black, linewidth=1pt, backgroundcolor=gray!10]
    \small

\textit{Annotation manual of test case generation task}:\\

We aim to modify the data within the solution's position to change the value in the cells of answer position. The steps are as follows:\\

1. Copy the original Excel file of the problem.\\
2. Locate the input position for the answer and the solution's position in the Excel summary sheet.\\
3. By understanding the question posed, you need to modify the data within the solution's position until the value in the answer cell changes. Record the location and sheet name of the changed cell in the Excel summary sheet (e.g., 'Sheet1'!A10). During modification, you may simulate corner case, such as deleting the cell's value, to mimic potential real-world incidents.\\
4. At this point, we obtain a new Excel file. Compared to the original Excel file, both the content of the table (i.e., data within the solution's position) and the values in the cells of answer position have changed.\\

For each data set, you need to modify it twice following the above steps, resulting in three pairs of test data (including the data from Task 1), named: 1\_{id}\_input.xlsx, 1\_{id}\_answer.xlsx, 2\_{id}\_input.xlsx, 2\_{id}\_answer.xlsx, 3\_{id}\_input.xlsx, 3\_{id}\_answer.xlsx.\\

Again, you have to make sure that the values in the cells that correspond to the solution's location in each of the three answer files are different.\\
- For the first modification: Ensure that at least one cell's value within the solution's position changes. Record the location and sheet name of one of the changed answer cells in the Excel summary sheet (e.g., 'Sheet1'!A10).\\
- For the second modification: Ensure that at least three cells' values within the solution's position change (if the solution's position contains fewer than three cells, change the value of just one cell). Record the locations and sheet names of three changed answer cells in the Excel summary sheet, separating the cell locations with spaces after the exclamation mark (e.g., 'Sheet1'!A10 A11 A12).\\

    \end{mdframed}
    \caption{The annotation manual of test case generation task.}
    \label{tab:test_case_gen}
\end{figure*}

%% file: table/prompt_single.tex
\begin{figure*}
    \centering
    \begin{mdframed}[linecolor=black, linewidth=1pt, backgroundcolor=gray!10]
    \small
You are a spreadsheet expert who can manipulate spreadsheets through Python code. \\

You need to solve the given spreadsheet manipulation question, which contains six types of information: \\
- instruction: The question about spreadsheet manipulation. \\
- spreadsheet\_path: The path of the spreadsheet file you need to manipulate. \\
- spreadsheet\_content: The first few rows of the content of speadsheet file. \\
- instruction\_type: There are two values (Cell-Level Manipulation, Sheet-Level Manipulation) used to indicate whether the answer to this question applies only to specific cells or to the entire worksheet. \\
- answer\_position: The position need to be modified or filled. For Cell-Level Manipulation questions, this field is filled with the cell position; for Sheet-Level Manipulation, it is the maximum range of cells you need to modify. You only need to modify or fill in values within the cell range specified by answer\_position. \\
- output\_path: You need to generate the modified spreadsheet file in this new path. \\

Below is the spreadsheet manipulation question you need to solve: \\
\#\#\# instruction \\
\{instruction\} \\
\\
\#\#\# spreadsheet\_path \\
\{spreadsheet\_path\} \\
\\
\#\#\# spreadsheet\_content \\
\{spreadsheet\_content\} \\
\\
\#\#\# instruction\_type \\
\{instruction\_type\} \\
\\
\#\#\# answer\_position \\
\{answer\_position\} \\
\\
\#\#\# output\_path \\
\{output\_path\} \\
\\
You should generate Python code for the final solution of the question.
    \end{mdframed}
    \caption{The prompt format used in single round setting.}
    \label{tab:single_prompt}
\end{figure*}

%% file: table/prompt_multiple.tex
\begin{figure*}
    \centering
    \begin{mdframed}[linecolor=black, linewidth=1pt, backgroundcolor=gray!10]
    \small
You are a spreadsheet expert who can manipulate spreadsheets through Python code. \\

You need to solve the given spreadsheet manipulation question, which contains six types of information: \\
- instruction: The question about spreadsheet manipulation. \\
- spreadsheet\_path: The path of the spreadsheet file you need to manipulate. \\
- spreadsheet\_content: The first few rows of the content of speadsheet file. \\
- instruction\_type: There are two values (Cell-Level Manipulation, Sheet-Level Manipulation) used to indicate whether the answer to this question applies only to specific cells or to the entire worksheet. \\
- answer\_position: The position need to be modified or filled. For Cell-Level Manipulation questions, this field is filled with the cell position; for Sheet-Level Manipulation, it is the maximum range of cells you need to modify. You only need to modify or fill in values within the cell range specified by answer\_position. \\
- output\_path: You need to generate the modified spreadsheet file in this new path. \\

Below is the spreadsheet manipulation question you need to solve: \\
\#\#\# instruction \\
\{instruction\} \\
\\
\#\#\# spreadsheet\_path \\
\{spreadsheet\_path\} \\
\\
\#\#\# spreadsheet\_content \\
\{spreadsheet\_content\} \\
\\
\#\#\# instruction\_type \\
\{instruction\_type\} \\
\\
\#\#\# answer\_position \\
\{answer\_position\} \\
\\
\#\#\# output\_path \\
\{output\_path\} \\
\\
The solution of the question can be generate through a multi-turn interaction and you can do two types of actions. \\
1. Spreadsheet information acquisition: You can generate Python code to obtain the information in the spreadsheet file. In the next turn, the execution result of you Python code will provide to you. \\
2. Question solution generation: You can generate Python code for the final solution of the question. If error occur when executing code, the error traceback will provide to you for code refinement.
    \end{mdframed}
    \caption{The prompt format used in multiple round setting.}
    \label{tab:multiple_prompt}
\end{figure*}